# Lesioned Multimodal Language Models Reproduce Aphasic Picture-Naming Patterns

Yong Yang[1,*], Xiang Guan[1], Sophie Arheix-Parras[4], Saeed Ahmadi[2], Roger Newman-Norlund[3,4], Leonardo Bonilha[5], Christopher Rorden[4], Julius Fridriksson[2,3], Rutvik H. Desai[4], Srihari Nelakuditi[1]

[1]*Department of Computer Science and Engineering, University of South Carolina;* [2]*Department of Communication Sciences and Disorders, University of South Carolina;* [3]*ALLT.AI, LLC;* [4]*Department of Psychology, University of South Carolina;* [5]*Department of Neurology, School of Medicine, University of South Carolina*

**Corresponding author: yongy@email.sc.edu*

## Abstract

Aphasia following stroke commonly produces systematic naming errors with characteristic profiles, but whether general-purpose language models not designed for clinical simulation can reproduce these patterns remains untested. We investigated (1) whether lesions or controlled perturbations to a multimodal language model can reproduce different types of errors in picture naming, and (2) whether the framework can reproduce the complete error profile of individual persons with aphasia (PWAs). Using LLaVA 1.6, we evaluated perturbation configurations that varied the layer, proportion, and amount of noise applied to model units. We examined 278 PWAs on the Philadelphia Naming Test, classifying responses into seven categories using a validated neural classifier. Six of seven response categories (correct, semantic, mixed, unrelated, neologism, no response errors) emerged at clinically-comparable proportions across distinct parameter space regions, with formal paraphasia being the exception. Searching the perturbation space revealed configurations that reproduced the individual error profile in at least six of seven categories for 97.8% of PWAs and in all seven categories for 79.5% of PWAs. Monte Carlo baselines confirmed that this matching reflects joint inter-category structure rather than marginal overlap. These results establish a quantitative framework for reproducing individual aphasic error patterns in picture naming. They suggest the potential for language models to serve as digital twins of individuals with post-stroke aphasia.

## Significance Statement

While artificial intelligence language models exhibit a wide range of capabilities, it is not clear whether they can serve as analogs of language processing in the human brain. Here, we tested whether ablations to language models can produce deficits that are similar to those found in post-stroke aphasia, using a picture naming task administered to a large cohort of persons with aphasia with varied error profiles. We demonstrate that controlled perturbations to a general-purpose multimodal language model reproduce the same error patterns observed in stroke survivors, with different regions of the lesion space associated with different error types. Formal paraphasias are a partial exception, arising in the model at substantially lower rates than in the patients whose error profiles they dominate. Moreover, we show that without any individual fine-tuning, specific perturbations to the model can produce complete error profiles of individual stroke survivors. The work establishes a quantitative framework for computationally characterizing individual aphasic picture-naming profiles. It demonstrates the potential of language models not just as powerful computational tools, but to serve as cognitive models of language processing in the human brain, with possible applications in testing and development of therapies in aphasia.

## Introduction

Aphasia following stroke produces systematic rather than random error patterns, with specific lesion characteristics yielding predictable failure modes (Fridriksson et al., 2022). These characteristic breakdown patterns have been extensively documented through standardized assessments. Earlier computational efforts to characterize aphasic naming errors have used purpose-built connectionist architectures with task- and person-specific parameter fitting (Dell & O'Seaghdha, 1992; Dell et al., 1997; Walker & Hickok, 2016). While language models exhibit a wide range of impressive capabilities, they are trained on vast amount of data that are not similar to those experienced any single individual, and the developmental trajectories of humans and language models also differ significantly. Hence, it is not clear whether language models can serve as cognitive analogs of language processing in the human brain.

Here, we address the question of whether a general-purpose multimodal language model, not designed for clinical simulation or as a cognitive model, can exhibit output-space error distributions that match those of individual PWAs when subjected to controlled perturbations of its internal representations, in the context of a standardized picture naming task.

This question has two distinct components:

1. Can controlled perturbations to a large language model reproduce each type of naming error characteristic of aphasia, including correct responses, semantic errors, formal errors, mixed errors, unrelated errors, neologisms, and no response?
2. Can the perturbation framework reproduce the complete error pattern (the joint distribution across all seven categories) of individual persons with aphasia?

The first question asks whether the model's perturbation space is sufficiently expressive for generating each clinical error category. The second asks whether this expressiveness extends to reproducing the specific combinations of error types that characterize individual PWAs, a far more stringent test, since it requires matching the joint distribution rather than marginal counts alone.

Modern multimodal language models are trained on broad language data without task-specific clinical modeling. If perturbations to such a model produce output-space error distributions that match individual PWA profiles (without task-specific design or per-patient parameter fitting) this would demonstrate that the perturbation space of a general-purpose model contains trajectories that reach the same regions of the error space occupied by clinical profiles. Recently, Wang et al. (2025, 2026) have shown that lesioning functionally specialized experts in Mixture-of-Experts language models can simulate the broad linguistic profiles of two aphasia subtypes, validated against clinical speech samples and the Western Aphasia Battery. The present work differs in scope: rather than targeting two subtype-level profiles, we characterize matching at the level of 278 individual error profiles, each evaluated against the joint distribution across seven response categories found in picture naming.

To address these questions, we adopt the Philadelphia Naming Test (PNT; Roach et al., 1996) as the behavioral assay. The PNT is a standardized picture naming task with established error taxonomies that enables direct quantitative comparison between model outputs and clinical data from 278 PWAs. Picture naming is among the most widely used instruments in aphasia assessment and is part of all standardized batteries. This is due to the fact that it assesses anomia, which is among the most common symptoms of aphasia. It requires coordination among multiple components of the language system, including semantic processing, retrieval of the correct word from by choosing among competing alternatives at phonological and/or semantic levels, and word production. Due to this fact, errors types in picture naming provide rich insights into specific components of the language system that are compromised, allowing valuable clinical insights into deficits and potential therapies from an ostensibly ‘simple’ task.

# Methods

## 2.1 Participants

### 2.1.1 Participant Characteristics

This study examined error patterns in picture naming from 278 PWAs drawn from the C-STAR (Center for the Study of Aphasia Recovery) Patient Dataset, a cohort of 410 aphasia patients assessed at the University of South Carolina; the analysis cohort comprised the 278 patients with complete PNT data. All participants were assessed using standardized aphasia batteries, including the Western Aphasia Battery-Revised (WAB-R; Kertesz, 2007). For 213 participants with complete WAB subscores, we computed the Aphasia Quotient (AQ) as a composite measure of aphasia severity (range: 0–100), with higher scores indicating better language function.

Inclusion criteria comprised: confirmed left-hemisphere stroke with resulting language impairment, completion of the full 175-item PNT with expert classification of all responses into standard error categories, and available demographic and clinical data. The dataset encompasses participants across the full severity spectrum, from very severe aphasia (AQ < 25) through mild impairment (AQ > 75).

To calculate category-specific tolerance thresholds for PWA matching, we utilized test-retest data from 156 participants in the PNT normative dataset who completed the PNT at two time points (B1 and B2 administrations). These paired measurements provided empirical estimates of natural within-participant variability for each error category, establishing statistically grounded matching criteria independent of the 278-participant matching cohort.

PNT responses are scored into seven categories consisting of **Correct** responses and six types of errors: **Semantic** (semantically related real word), **Formal** (phonologically similar real word), **Mixed** (both semantic and phonologically related), **Unrelated** (real word with neither relation), **Neologism** (non-lexical novel form), and **No Response** (failure to name, including empty outputs, descriptions, or explicit statements of inability). Formal definitions, illustrative examples (Table 2), and the corresponding classification decision tree are given in Section 2.3.

Table 1 summarizes the distribution of PNT response counts across the seven categories, reported as the mean, standard deviation, and range over the 278 PWAs. For participant demographics, WAB-AQ averaged 63.4 (SD 24.9, range 13.3–99.6) and was available for 213 of the 278 participants, recomputed from Western Aphasia Battery subscores; the remaining 65 lacked complete WAB subtest scores but had full PNT response data for the matching analysis. Age averaged 62.30 years (SD 14.12, range 21–80), and time post-stroke averaged 3.93 years (SD 4.22, range 0.37–19.48).

| Response Category | Mean | SD | Range |
|---|---|---|---|
| Correct | 82.7 | 64.6 | 0–175 |
| Semantic | 7.3 | 9.6 | 0–72 |
| Unrelated | 10.7 | 20.0 | 0–140 |
| Formal | 20.3 | 25.7 | 0–128 |
| Mixed | 1.6 | 2.2 | 0–18 |
| Neologism | 14.0 | 24.4 | 0–157 |
| No Response | 38.4 | 46.5 | 0–175 |

*Table 1: PNT response-category distribution (N = 278). Values are the mean, standard deviation, and range of response counts, out of 175 scorable items, across the cohort; each participant's seven categories sum to 175.*

### 2.1.2 PNT Response Patterns

Response patterns varied systematically with aphasia severity. Less-impaired participants produced predominantly correct responses, whereas the most severely impaired produced predominantly no-response errors with minimal correct naming. Formal errors occurred across all severity levels, and neologisms followed a clear severity gradient, increasing with impairment. The cohort therefore spans the full range of clinical severity, providing the variation required for the individual-level matching analysis that follows.

## 2.2 Materials

The PNT has been extensively used in aphasia research and is the standard instrument for quantifying naming error distributions in PWAs (Roach et al., 1996; Mirman et al., 2010; Mirman & Britt, 2014). Its standardized administration and quantitative error taxonomy have enabled systematic comparison of error patterns across populations and across computational characterizations of those patterns (Walker et al., 2018). Picture-naming tests such as the PNT (and related instruments such as the Boston Naming Test) are, moreover, frequently used as primary outcome measures in clinical trials of aphasia therapy.

**Model Selection Rationale.** LLaVA 1.6 (Liu et al., 2023) was selected as the test model on three grounds. First, at the inception of the present study, LLaVA 1.6 was the most capable openly-available multimodal language model whose weights, training procedure, and architectural specification were fully accessible, enabling the unit-level perturbation protocol described in Section 2.3.1. Closed multimodal systems do not permit weight-level intervention. Second, the LLaVA 1.6 architecture is transparent for analyses of language-relevant computational mechanisms: a CLIP visual encoder feeds a vision-language projection module that injects fixed-length visual tokens into a Vicuna-13B language backbone, with no architectural devices (e.g., gated mixtures of experts, retrieval modules, tool-use heads) intervening between visual input and lexical output. This minimal structure provides a defensible baseline for asking which layers and unit subsets contribute to which error categories. Third, the multimodal capability is sufficient to support digital adaptations of standard psycholinguistic and cognitive assessments beyond the PNT, including object-naming, action-naming, and category-fluency tasks, providing a foundation on which the present picture-naming framework can be extended in future work.

## 2.3 Experimental Design

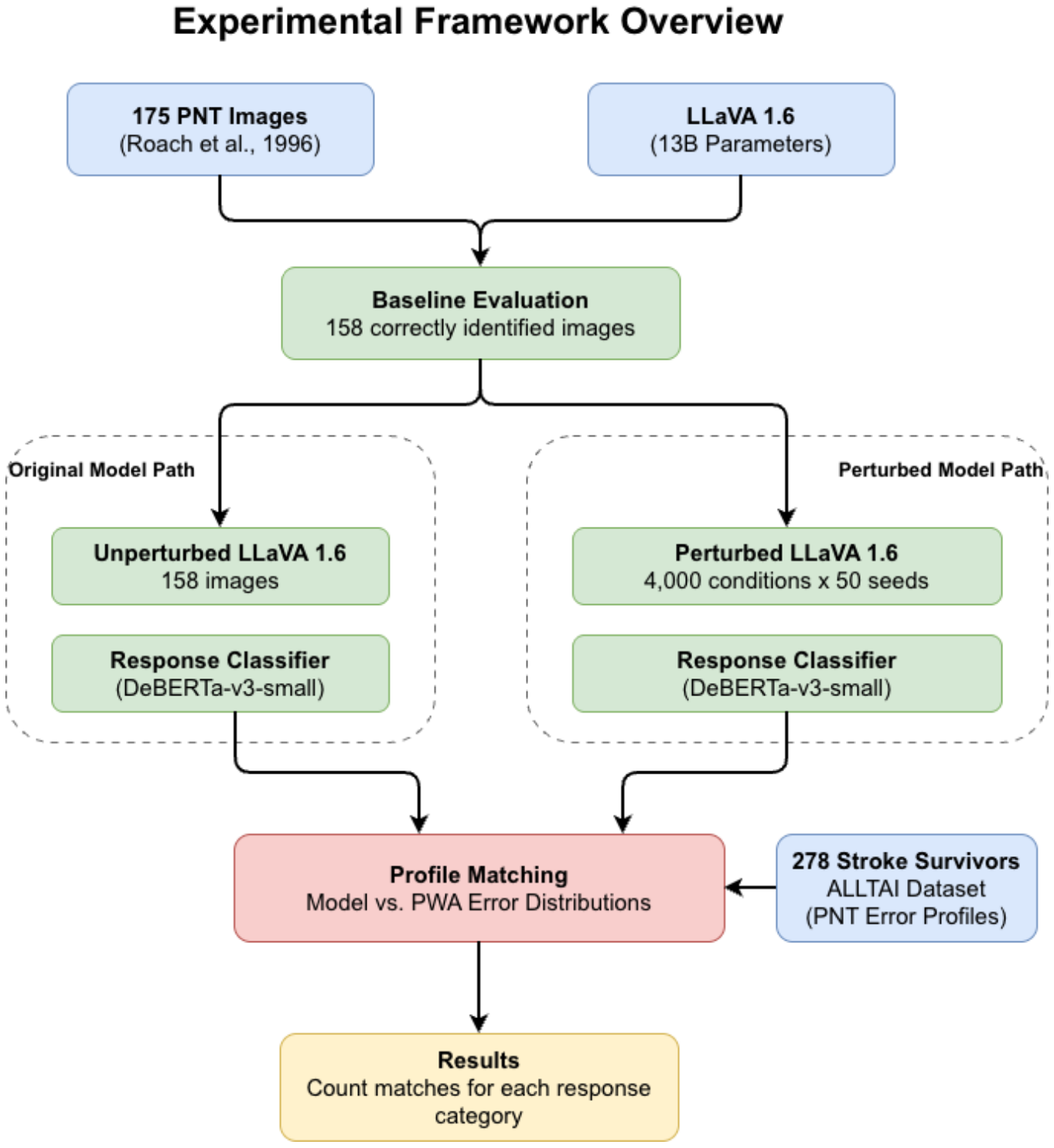

*Figure 1: Experimental Framework Overview. Processing of the standard 175-image PNT stimulus set through original and perturbed LLaVA 1.6 models, followed by classifier-based response categorization into seven error categories, and quantitative matching of model error distributions against 278 PWA profiles from the C-STAR Patient Dataset. All reported analyses use only the standard 175 PNT items.*

As shown in Figure 1, our experimental procedure follows a comparative analysis framework. We presented all 175 scorable PNT images (Roach et al., 1996) to the unperturbed model with the standardized prompt: "What is shown in this image? Provide a single word answer." This baseline evaluation identified 158 images for which the model produced correct responses, matching established target labels through exact string matching. To ensure that observed errors under perturbation conditions resulted from computational disruption rather than baseline model limitations, the 17 images the model failed to identify at baseline were excluded from perturbation experiments. The remaining 158 correctly identified images were then systematically evaluated under controlled perturbation conditions, with disruptions defined by three parameters: noise level, modification proportion, and target layer. To maintain comparability with the 175-item clinical PNT scoring standard, perturbation results from the 158-image subset were proportionally scaled to the full 175-item count for all subsequent analyses.

### 2.3.1 Perturbation Protocol

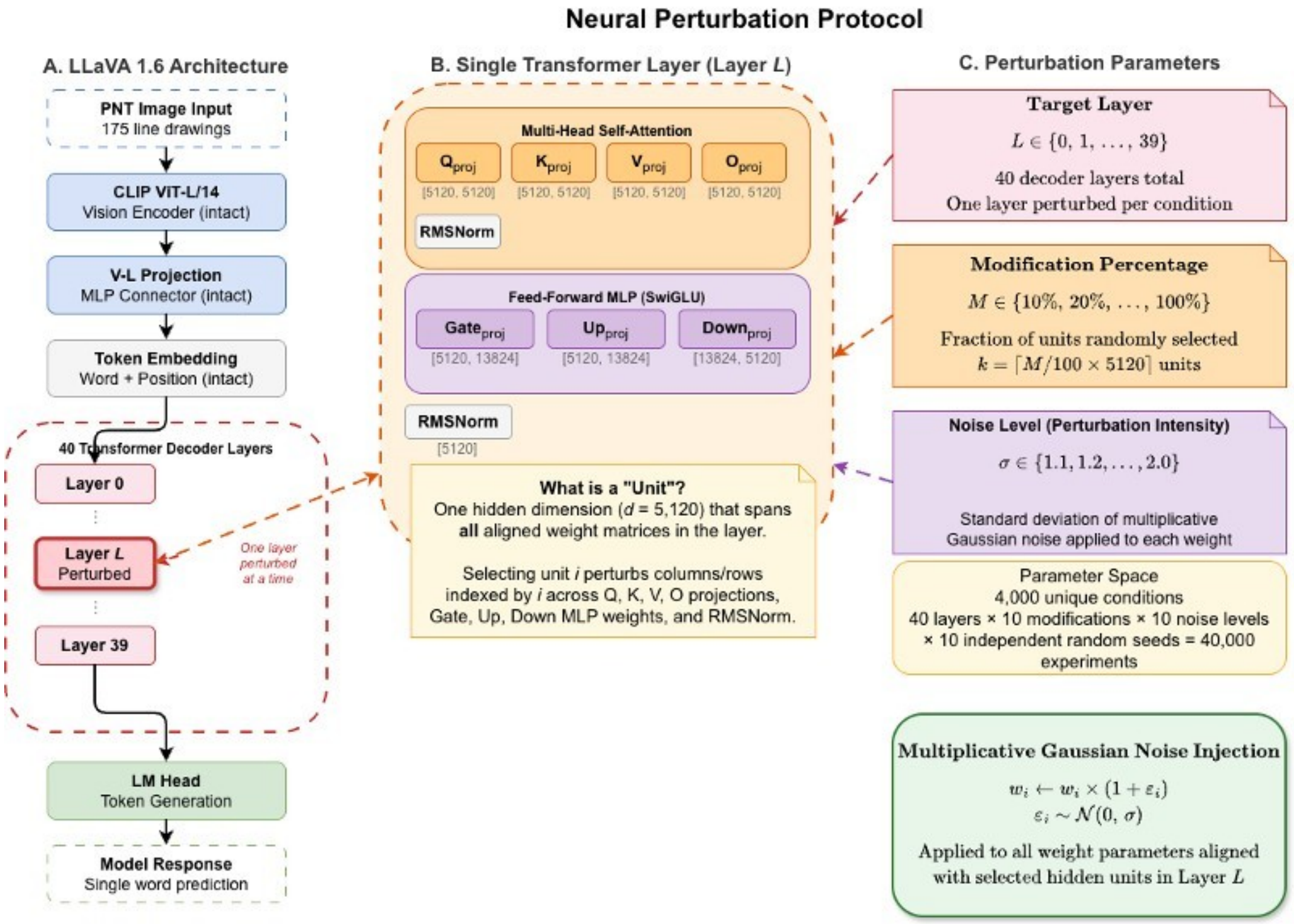

*Figure 2: Perturbation Protocol. (A) LLaVA 1.6 architecture with 40 transformer decoder layers (layers 0–39) as perturbation targets; vision encoder and projection remain intact. (B) Anatomy of a single layer showing all weight matrices aligned with the hidden dimension (d = 5,120). A "unit" corresponds to one hidden dimension spanning all aligned parameters. (C) Three perturbation parameters and the multiplicative Gaussian noise injection formula.*

As illustrated in Figure 2, our perturbation protocol targeted the 40 transformer decoder layers (layers 0–39) of the model. The vision encoder (CLIP ViT-L/14), vision-language projection module, token embedding layer, and language model head remained intact throughout all experiments. Only one layer was perturbed per experimental condition; this layer-localized intervention design follows methodological precedents in mechanistic interpretability of transformer models (Olah et al., 2020; Geva et al., 2022; Meng et al., 2022). This selective approach parallels a key feature of aphasia: focal damage to perisylvian language cortex disrupts naming while visual object recognition remains largely preserved (Goodglass et al., 2001; Hillis, 2007), just as our perturbations target the language model backbone while leaving the vision encoder intact.

Within each layer, the fundamental perturbation target is a "unit," defined as one dimension of the 5,120-dimensional hidden representation. Selecting a unit for perturbation affects all weight matrices aligned with that hidden dimension across the layer's sub-modules: the columns of Q, K, V, and O attention projections, the columns of Gate and Up MLP projections, the corresponding row of the Down MLP projection, and the associated RMSNorm parameter. This approach ensures that perturbation of a unit disrupts its contribution throughout the layer's computation, rather than targeting isolated weight elements.

The perturbation procedure was controlled by three systematically varied parameters (Figure 2C). **Target layer** (*l*) specified which single layer (0–39) received perturbations within each experimental condition. **Modification proportion** (*p*) determined the fraction of units within the selected layer that were randomly chosen for perturbation, ranging from 10% to 100% in 10% increments; specifically, $k = \lceil p/100 \times 5{,}120 \rceil$ units were selected per condition. **Noise level** (σ) controlled the intensity of perturbation applied to the selected units, using multiplicative Gaussian noise: each affected weight $w_i$ was replaced by $w_i \times (1 + \varepsilon_i)$, where $\varepsilon_i \sim N(0, \sigma^2)$, with σ (the standard deviation) ranging from 1.1 to 2.0 in 0.1 increments. This systematic parameterization created a three-dimensional space of 4,000 unique experimental conditions (40 layers × 10 modification proportions × 10 noise levels); each condition is a perturbation configuration, denoted by the triplet (l, p, σ). Each condition was applied to the complete set of 158 baseline-correct images across 50 independent random seeds, yielding 200,000 total experiments. Throughout, p denotes the modification proportion; statistical significance is reported in full as the p-value to avoid ambiguity with this symbol.

The experiments reported in the main text apply isotropic Gaussian weight noise uniformly across the entire 5,120-unit language modeling space at the target layer (complete-unit, CU, configuration); this is the default throughout. A complementary control restricted perturbation to a Localizer-defined subset of equally-sized language units (language unit, LU, configuration), with the subset computed by identifying the units whose activations were most differentially engaged during the PNT naming task. On a 20-seed shared subsample of the main analysis, the LU configuration yielded patient-aggregate matching equivalent to the CU configuration (Δ high-quality matching = −1.08 percentage points at the retest tolerance, well within seed-level variability), indicating that the matching reported here does not depend on the CU choice. Full methodological details for the LU control are provided in Supplementary Section S5; per-PWA results are available from the authors on request.

### 2.3.2 Response Classification for Aphasia Error Types

Our classification system categorizes naming responses into seven error types following the standardized PNT scoring taxonomy (Roach et al., 1996; Mirman et al., 2010): Correct, Semantic, Formal, Mixed, Unrelated, Neologism, and No Response. To ensure methodological consistency between large language model (LLM)-generated responses and PWA clinical data, we employed a fine-tuned neural network classifier validated against expert speech-language pathology (SLP) annotations.

**Classification Model Architecture.** We developed a feature-augmented neural classifier built on the DeBERTa-v3-small backbone (He et al., 2021), fine-tuned on 3,160 expert-annotated naming responses. The classifier receives feature-augmented input combining target-response word pairs with six computed binary features: exact match (including singular/plural variants), phonological relatedness (a binary flag set when the target and response share an onset phoneme, a word-final phoneme, the primary-stressed vowel, or two or more phonemes, derived from CMU Pronouncing Dictionary transcriptions), onset phoneme match, word-final phoneme match, lexicality (presence in the CMU Pronouncing Dictionary), and response length flag. This feature augmentation provides explicit phonological and lexical signals that guide classification decisions. For out-of-vocabulary strings without a dictionary pronunciation, the re-scoring analysis in Supplementary Section S6 computes phonological relatedness from grapheme-to-phoneme transcriptions and assesses lexicality by WordNet membership, which excludes proper nouns.

**Validation Against Expert Annotations.** The response classifier achieved 89.21% accuracy and Cohen's Kappa of 0.87(Almost Perfect agreement, per Landis & Koch, 1977) against expert SLP annotations (see Supplementary Material S4 for comprehensive validation). The classifier achieves F1 = 0.875 for No Response detection, a category where rule-based methods fail entirely (F1 = 0.000). This capability is essential because No Response errors constitute 11.6% of expert-annotated data and 15.3% of PWA clinical data.

**Methodological Consistency.** Distribution-level validation confirmed that classifier outputs closely match PWA data patterns (Pearson r = 0.9757, Jensen-Shannon divergence = 0.0000), approaching the alignment observed between expert annotations and PWA data (r = 0.9779). This consistency ensures that cross-population comparisons between LLM perturbation errors and PWA naming errors reflect genuine similarities rather than classification artifacts.

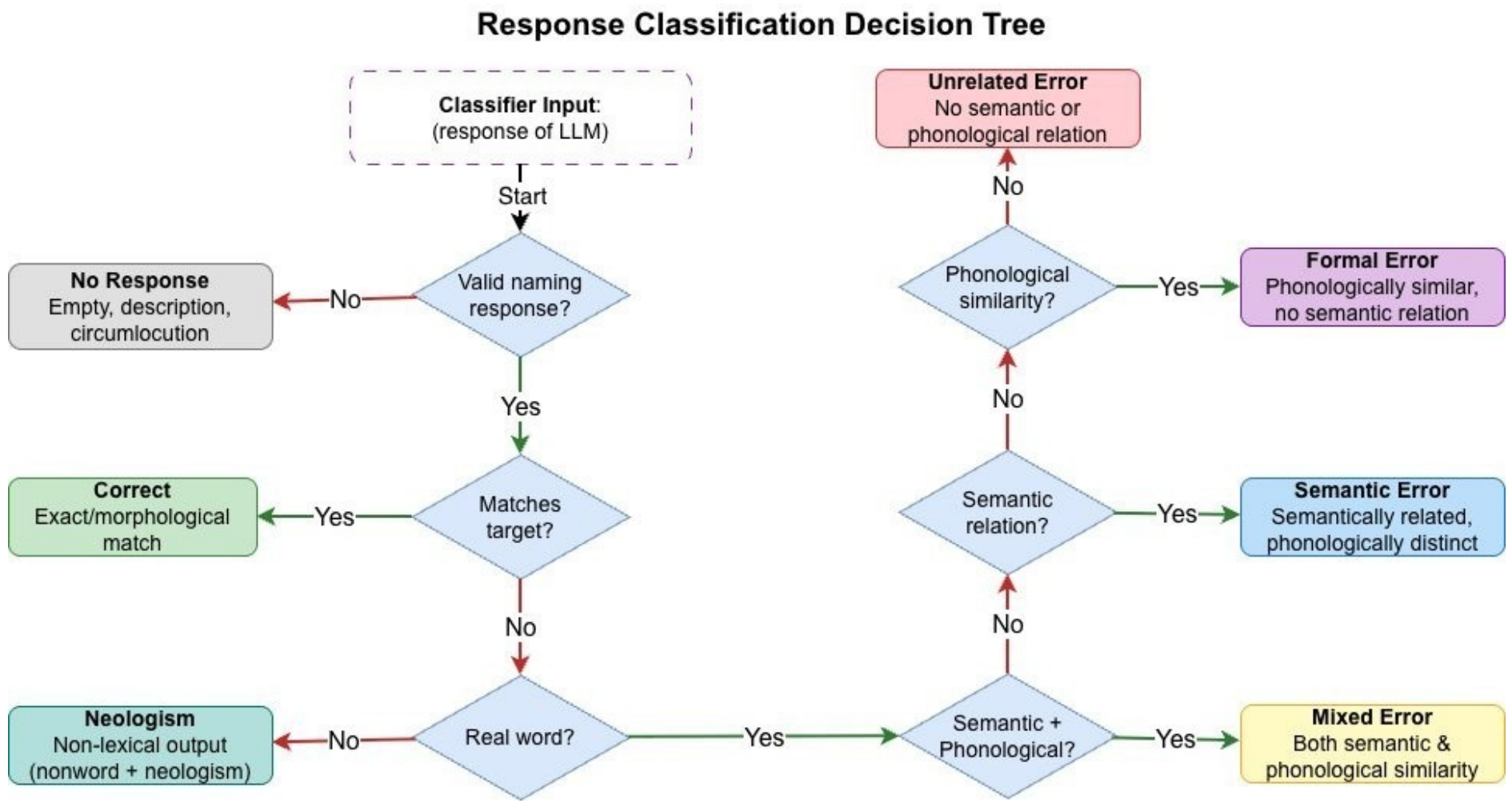

*Figure 3: Response Classification Decision Tree. Hierarchical classification framework for categorizing model outputs into seven response types (Correct, Semantic, Unrelated, Formal, Mixed, Neologism, No Response) based on established aphasia error taxonomies. Classification employs a validated feature-augmented neural classifier built on the DeBERTa-v3-small backbone.*

The seven response categories are defined as follows (see Figure 3 for the classification decision tree). These rules were designed to mirror, as closely as possible, the rules used by SLPs for human responses. (1) **No Response**: the model fails to generate a valid naming response, including empty outputs, descriptions, circumlocutions, or explicit statements of inability to name. (2) **Correct**: the response matches the target word, including morphological variants. (3) **Semantic Error**: the response is semantically related to the target but phonologically unrelated (e.g., "moth" for "butterfly"). (4) **Formal Error**: the response shares phonological similarity with the target but lacks a semantic connection (e.g., "buttercup" for "butterfly"). (5) **Mixed Error**: the response exhibits both semantic and phonological similarity to the target (e.g., "dragonfly" for "butterfly"). (6) **Neologism**: the response is a novel or nonexistent word (e.g., "flitterfly" for "butterfly"). The standard PNT scoring system distinguishes two subtypes: phonologically-related nonwords, which share phonemes with the target, and abstruse neologisms, which do not (Roach et al., 1996). We consolidate the two subtypes into a single non-lexical category for the interactive-two-step model-fitting analysis, following standard practice in that framework (Foygel & Dell, 2000; Schwartz et al., 2006). A re-scored split of the two subtypes by phonological relatedness to the target, using the standard PNT phonological-similarity criterion, is reported in Supplementary Section S6. (7) **Unrelated Error**: the response is a real word bearing neither semantic nor phonological relationship to the target (e.g., "table" for "butterfly").

### 2.4 Statistical Methods

Statistical analyses employed comprehensive hypothesis testing protocols with rigorous baseline validation. The empirically validated random baseline was established through Monte Carlo simulations using dual randomness control frameworks, as detailed in Supplementary Material S2. We tested the null hypothesis that the mean proportion of high-quality PWA matches (≥6 categories) equals the empirically validated baseline against the alternative hypothesis of significant difference, using two-tailed one-sample t-tests with cross-seed validation.

A Bonferroni-corrected significance threshold ($\alpha = 0.05/3 \approx 0.0167$, across the three primary tests: chi-square goodness-of-fit, one-sample t-test, and cross-seed ANOVA), effect size calculations (Cohen’s h and d (Cohen, 1988)), and confidence interval estimations were conducted following established protocols detailed in Supplementary Material S2. The result is the per-PWA maximum (Section 2.5): the condition attaining the highest matched-category count. The effect of a given (l, p, σ) configuration, represented by its median across seeds, is reported in Supplementary Material S3. All analyses used Python 3.12.3 with SciPy 1.15.3, employing fixed random seeds for reproducibility while maintaining independent experimental seeds for stability assessment.

### 2.5 Data Analysis

We systematically examined relationships between perturbation parameters (noise level, modification proportion, and target layer) and resulting error distributions through complementary analytical approaches. Parameter space visualization included: (1) error category distributions across noise levels (1.1–2.0σ) and modification proportions (10–100%), (2) layer-specific error patterns across all 40 layers, and (3) individual PWA profile matching optimization.

Additionally, we examined the general characteristics of the ($l$, $p$, σ) triplet, collapsing across random seeds, by reporting the median category matches for the triplet with the optimum match.

### 2.6 Data Availability

The input data used in this study is drawn from the C-STAR (Center for the Study of Aphasia Recovery) Patient Dataset. The data supporting the findings of this study are available from the corresponding author upon reasonable request.

## Results

### 3.1 Reproduction of Individual Error Categories Under Controlled Perturbations

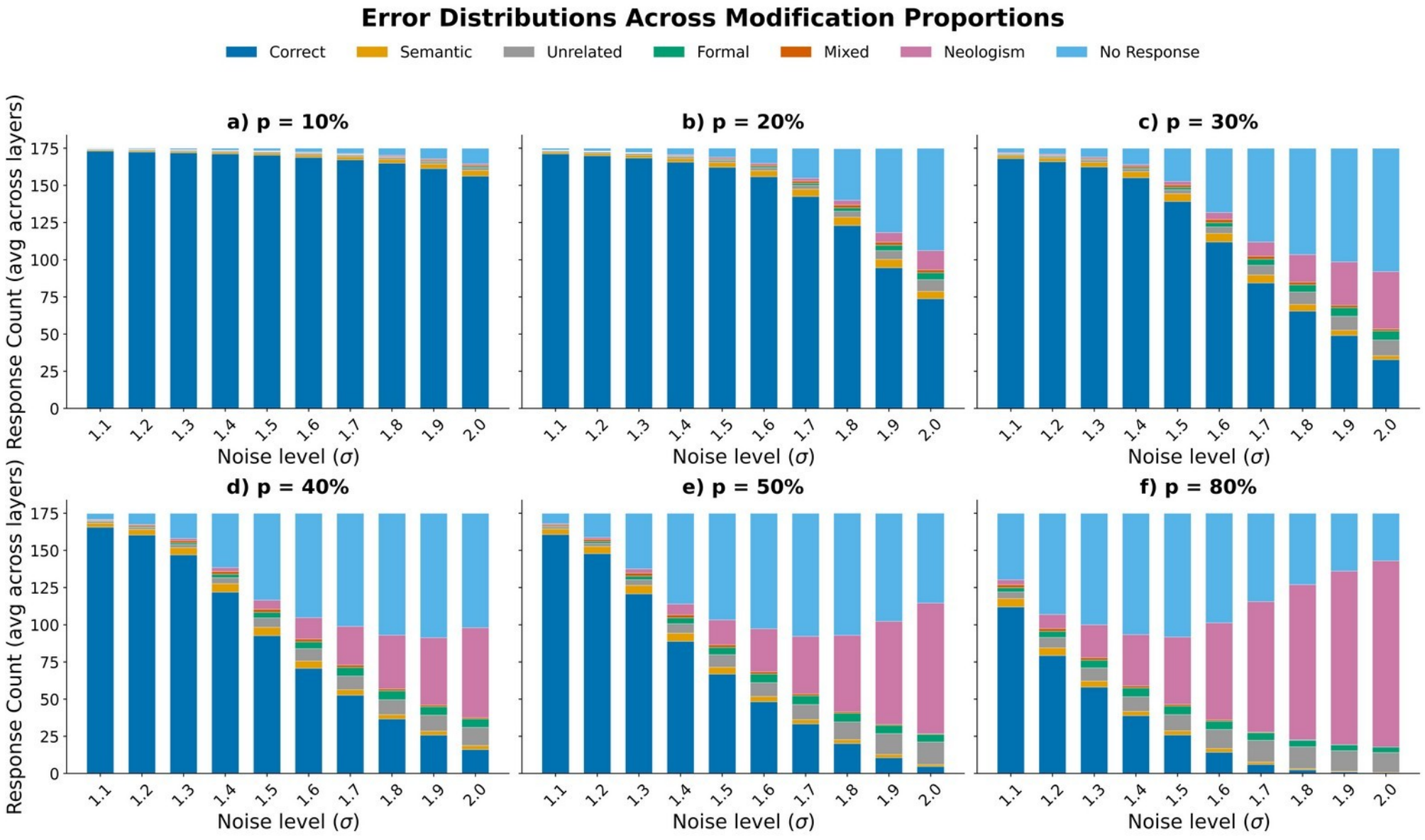

*Figure 4: Error Distributions Across Modification Proportions. Comparison of error type distributions at 10%, 20%, 30%, 40%, 50%, and 80% modification proportions, plotted against noise level (x-axis, 1.1–2.0σ) and averaged across all 40 layers and 50 random seeds; bars are stacked by the seven response categories. Y-axis represents response count scaled to the full 175 scorable PNT items.*

We systematically examined how controlled perturbations affect the error patterns produced by the multimodal language model, addressing our first question: whether perturbations can reproduce each type of naming error characteristic of aphasia. Figure 4 compares error distributions across noise levels at multiple modification proportions, averaging results across all 40 layers and 50 random seeds.

At 50% modification, the model demonstrated a gradual transition through distinct error categories as noise increased. Intermediate error types (particularly neologisms) appeared prominently at moderate noise levels before No Response errors dominated at higher perturbation intensities. In contrast, at 80% modification, correct responses declined more rapidly with fewer intermediate stages: higher modification proportions produced more abrupt transitions in the output mixture. Both conditions showed correct responses decreasing as noise increased, with No Responses eventually dominating at the highest perturbation intensities.

Representative examples of model output for each of the five non-trivial error categories are shown in Table 2; these are drawn from the inference output of perturbation runs at the configurations indicated. The correct and no-response categories are excluded from the table because the former is trivially "Target" exactly and the latter is by definition an empty output.

| Error Category | Target | Model Output | Configuration (l, p, σ) |
|---|---|---|---|
| **Semantic** | peas | corn | $l$=4, $p$=10%, σ=1.4 |
| **Unrelated** | tractor | Ideas | $l$=10, $p$=50%, σ=1.4 |
| **Formal** | pear | Pair | $l$=1, $p$=10%, σ=1.1 |
| **Mixed** | pen | Pencil | $l$=0, $p$=10%, σ=1.1 |
| **Neologism** | tractor | sassd | $l$=23, $p$=40%, σ=2.0 |

*Table 2: Representative examples of model output for each of the five non-trivial error categories under perturbation. Each row reports one PNT item from the inference set, the model's output under the indicated (l, p, σ) configuration, and the category assigned by the DeBERTa-v3-small feature-augmented classifier (Section 2.3.2). Examples were drawn from the inference outputs across the 50 seeds used in the main analysis, sampled to illustrate clinically recognizable instances of each category. Where possible the same target word (tractor) is used across categories to emphasize that the same input can yield different error types under different perturbation configurations; the formal- and semantic-error rows instead use targets (hammer and peas, respectively) for which the characteristic substitution is most transparent.*

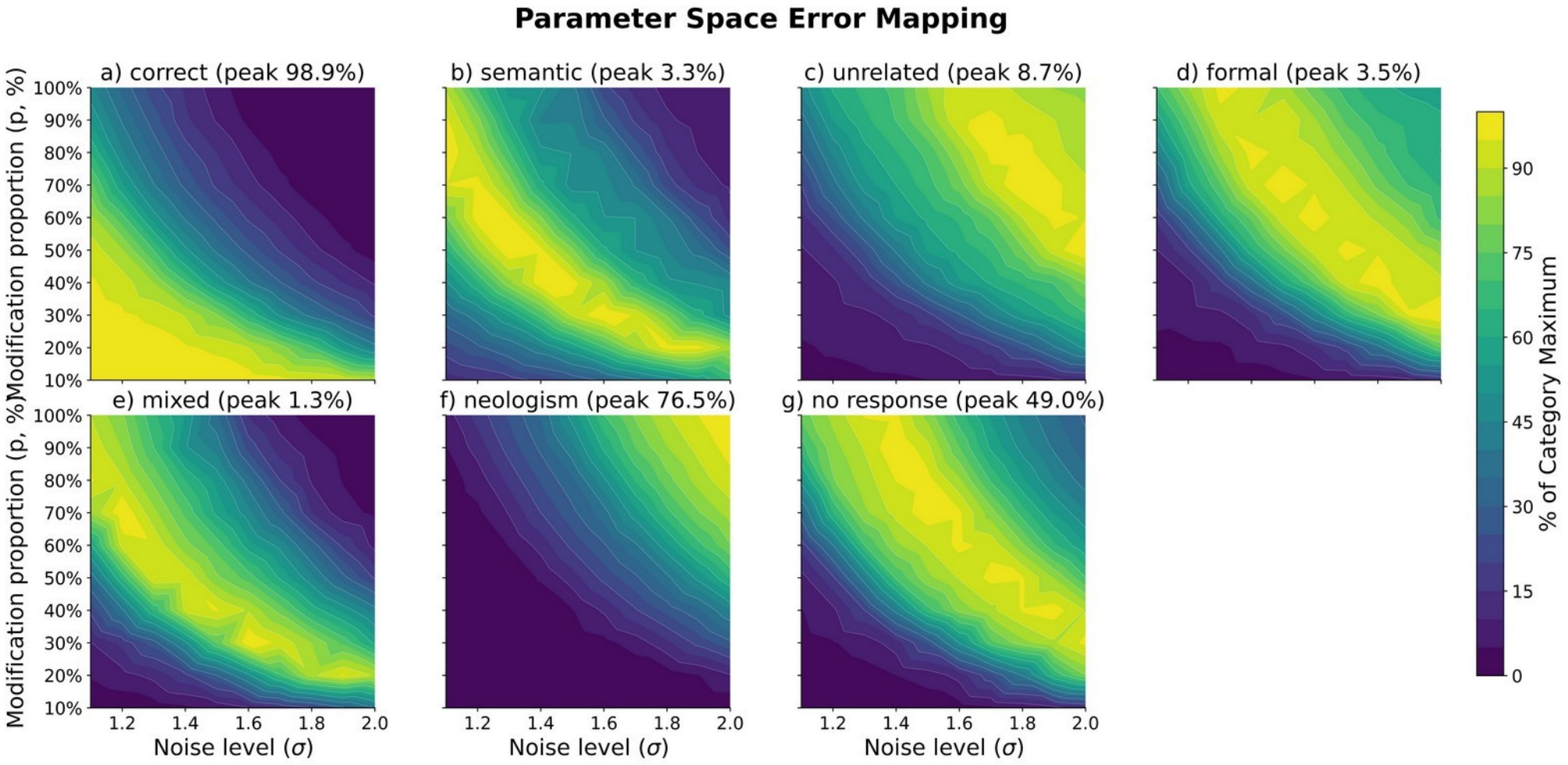

*Figure 5: Parameter Space Error Mapping. Heat maps showing the distribution of each of the seven response categories across noise levels (1.1–2.0σ, x-axis) and modification proportions (10–100%, y-axis), averaged across all 40 layers and 50 random seeds. The color scale in each panel is normalized to its own per-category maximum (noted in the subtitle of each panel), which differs substantially between categories.*

Figure 5 reveals distinct parameter regions where specific error categories dominate the output mixture. Correct responses concentrate in regions of low perturbation, forming a gradient that rapidly diminishes above 40% modification and 1.6σ noise. Semantic errors form a diagonal band from moderate to high perturbation levels. Neologisms appear in concentrated clusters at specific perturbation combinations, while No Response errors systematically increase with perturbation intensity, eventually dominating at the highest perturbation levels. Because each panel is averaged across all 40 layers, these peak percentages understate the per-condition magnitudes that the individual matching (Figures 7–8) draws on.

Against this picture of distinct parameter regions producing distinct error types, one boundary should be noted. The observed error patterns reflect a constraint on lexical retrieval rather than an absence of phonemic structure: genuine formal errors remain bounded at low magnitudes (≈ 3.5% of responses layer-averaged, Figure 5d) because selecting a phonologically similar real word requires retrieving a lexical neighbor that the model does not index by sound, whereas perturbation readily yields phonologically-related non-lexical forms (Supplementary Section S6). The model therefore produces partial, target-related phonological output but rarely resolves it onto a real word. Neologisms, by contrast, emerge when perturbations disrupt the token generation process, producing lexically invalid sequences; this failure mode is well within reach of the model's parameter space and dominates many parameter regions (Figure 5f, peak ≈ 77% of responses). This asymmetry is returned to in Discussion (Architectural Specialization as Convergent Prediction).

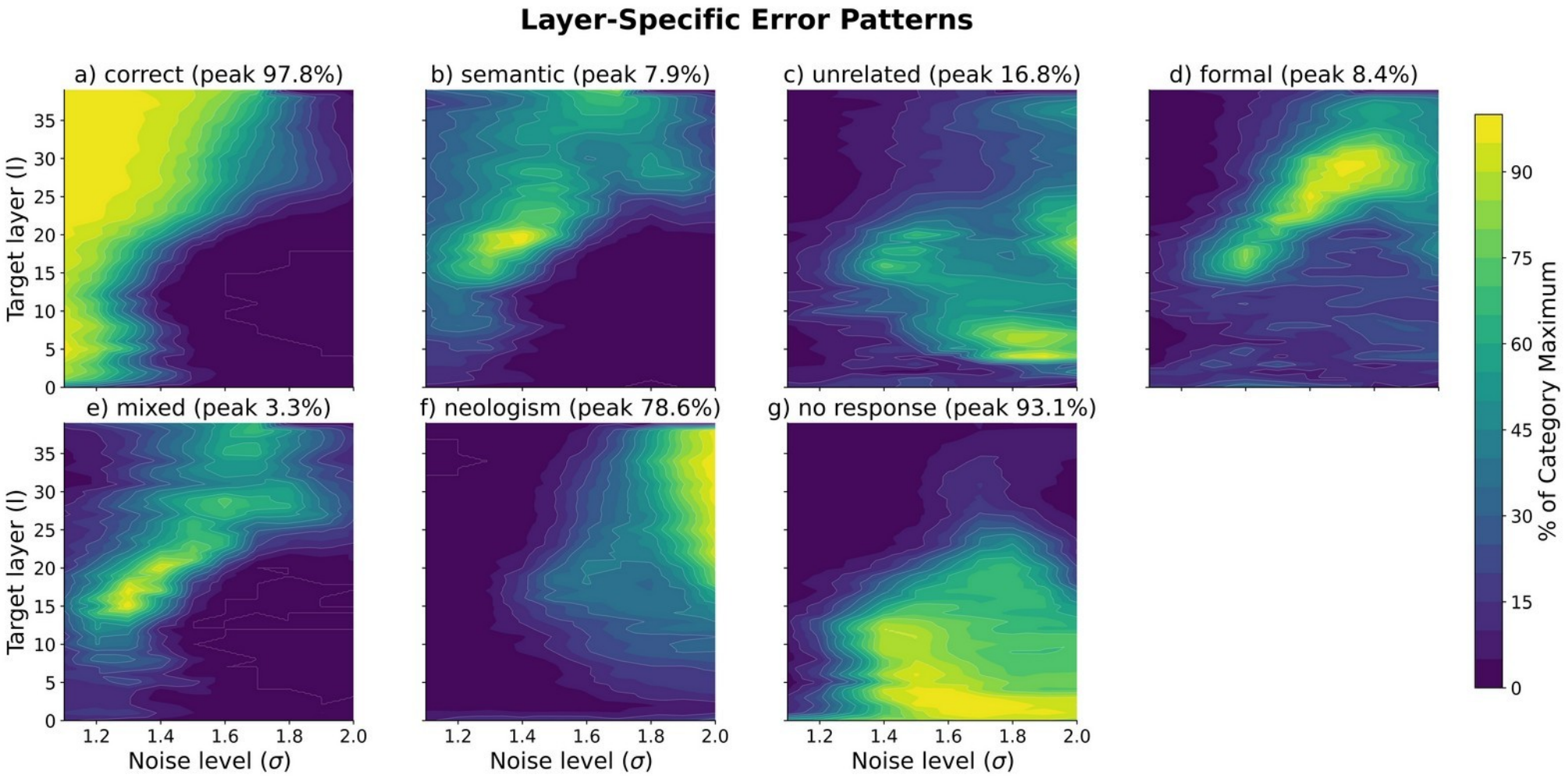

*Figure 6: Layer-Specific Error Patterns. Heat maps showing the distribution of each of the seven error categories across layers and noise levels at the 50% modification proportion, averaged across random seeds. As in Figure 5, color encodes each category's count as a percentage of its own per-panel maximum The panels reveal layer-specific vulnerability patterns that differ across error categories.*

Figure 6 shows how output-space error mixtures vary across the 40 layers at fixed modification (50%) and across noise levels. Correct responses persisted under perturbation of upper layers (25–39) even as noise increased, until degrading rapidly above 1.6σ. Perturbation of lower layers (0–9) produced primarily No Response outputs, with isolated regions of correct responses at specific noise levels. Semantic errors emerged primarily under perturbation of middle-to-upper layers (17–29); mixed errors distributed more broadly across layers 5–33; neologisms concentrated in the upper layers (≈28–39) at high noise levels. No Response errors showed a clear gradient from lower to higher layers as noise increased. These are output-space descriptions of how the layer-of-perturbation parameter shapes the error mixture.

Together, these results demonstrate that language model breakdown under perturbation is structured rather than random, with distinct parameter regions producing distinct error profiles. Six of the seven clinical error categories (correct, semantic, unrelated, mixed, neologism, and no-response) emerge across the parameter space at clinically-comparable proportions, each producible in the magnitudes observed across the PWA cohort. The seventh category, formal paraphasias, appears in the model's output but is systematically under-produced relative to the clinical range: model formal errors peak at ≈ 3.5% of responses in the layer-averaged view (Figure 5d) and ≈ 8.4% at the single most susceptible layer (Figure 6d), whereas formal-paraphasia-dominant PWAs exhibit formal error proportions averaging 67 of 175 responses (≈ 38%). The first question is therefore answered affirmatively for six of seven categories at magnitude correspondence and for all seven at existence; the systematic under-production of formal errors is an architectural feature of LLaVA's token-level vocabulary, returned to in Discussion (Architectural Specialization as Convergent Prediction).

Quantitative dose-response analysis across all 40 layers and 50 seeds further characterizes this structured degradation. Correct responses declined monotonically from 96.5% at the lowest perturbation severity to 0.6% at the highest, with the 50% accuracy threshold occurring at a composite severity midpoint of 0.73 (on the noise × modification scale). Error type transitions followed an orderly sequence: neologisms first exceeded correct responses at severity 1.21, while no-response errors peaked at intermediate severities before declining as neologism production increased, a pattern consistent with graded rather than abrupt degradation. The dose-response structure was layer-dependent: perturbation of late layers (30–39) produced a gradual correct → semantic → neologism transition (with formal paraphasias absent from this transition, as discussed under Architectural Specialization below), whereas perturbation of early layers (0–9) produced rapid collapse to neologism and no-response outputs at lower severity thresholds. The mean accuracy at layer 0 (20.2%) was 3.6 times lower than at layer 39 (71.9%), confirming a monotonic accuracy gradient across the model's layers. Our second question, regarding whether this output-space structure is rich enough to reproduce the joint error distributions of individual PWAs, is tested in the matching analysis that follows.

## 3.2 PWA Profile Matching

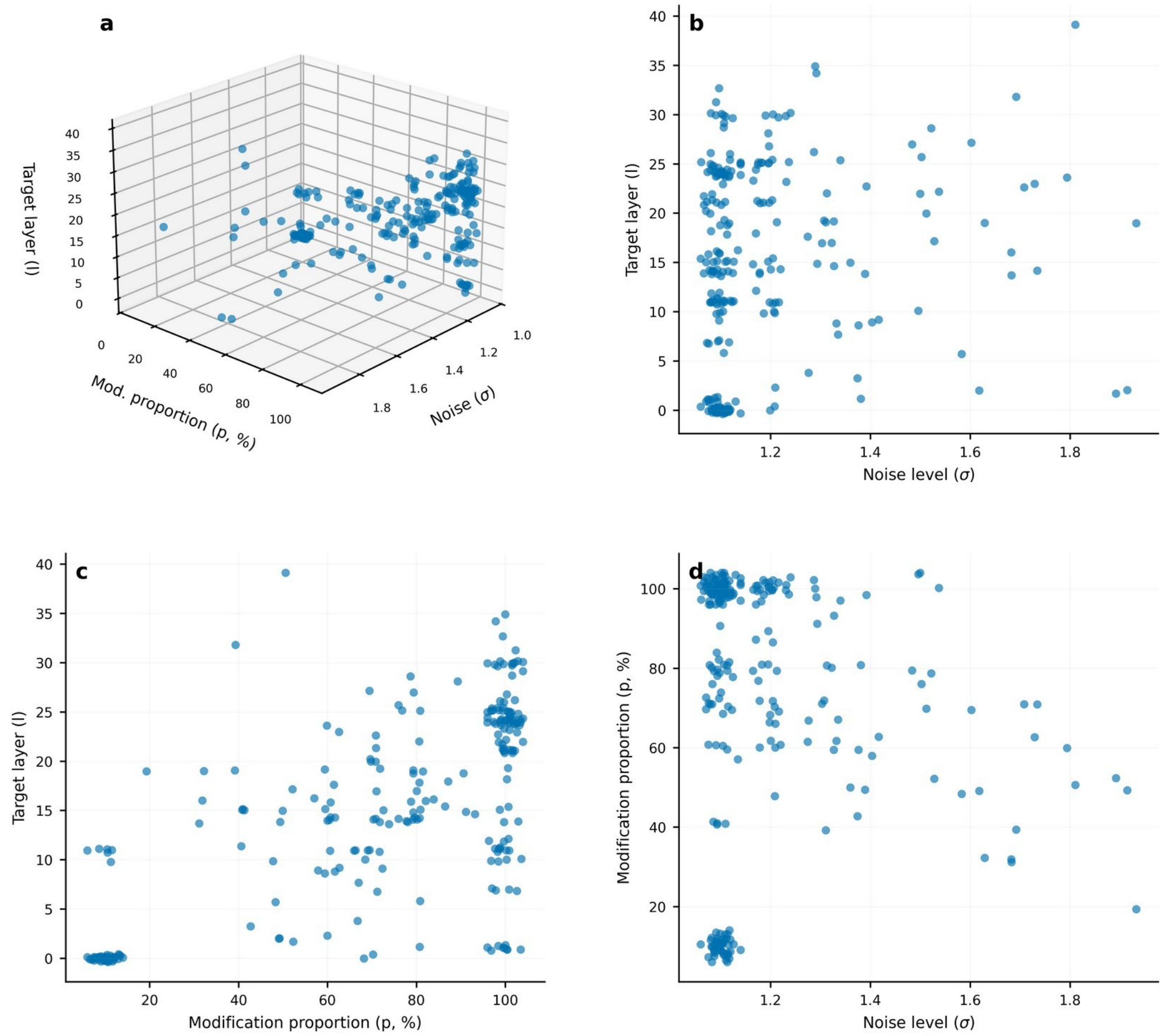

*Figure 7A: Best-fit perturbation configurations for individually matched patients, shown in the (**l**, **p**, σ) parameter space for a single representative seed. Each marker is one person with aphasia (PWA) whose best-fit configuration reproduced the individual profile in at least six of seven categories; for this representative seed, 255 of 278 PWAs (91.7%) met this high-quality criterion, comparable to the cross-seed average (90.9%). (a) the joint three-dimensional distribution of layer, modification proportion, and noise level (σ); (b–d) its two-dimensional projections: (b) layer versus noise, (c) layer versus modification, and (d) modification versus noise, in which mean modification rises monotonically with mean noise (panel d). The configurations occupy a structured, severity-graded region (more impaired profiles are matched at higher modification and noise), consistent with the population-level parameter organization in Figures 5 and 6.*

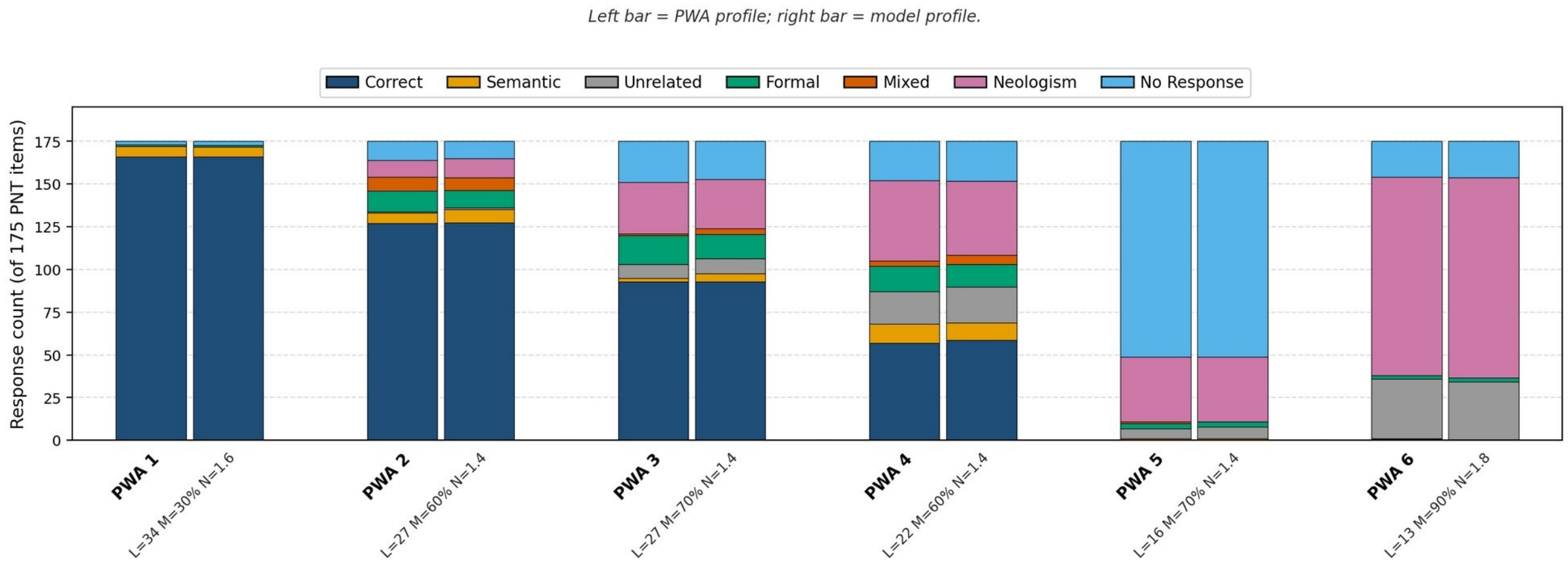

*Figure 7B: PWA–model error-distribution comparisons for six representative stroke survivors spanning the error-type and severity range, from no-response– and neologism-dominant profiles to correct-dominant profiles. Each PWA is shown as a pair of adjacent stacked bars: the left bar is the patient's observed response distribution and the right bar is the model output at that PWA's best-fit (**l**, **p**, σ) configuration, indicated beneath each pair.*

Having established that perturbation-induced errors are structured and span all seven clinical categories, we next asked whether this structure is rich enough to reproduce individual clinical profiles, the second of our two central questions. Consistent with standard practice in computational models of aphasic naming (Foygel & Dell, 2000; Schwartz et al., 2006; Dell et al., 2013;

Walker & Hickok, 2016), we report the result, the per-PWA maximum-matching configuration, defined in Section 2.5; the median across seeds for that configuration is provided in Supplementary Material S3.

Searching the 4,000-condition space across 50 seeds, a configuration reproduced the individual profile within the category-specific retest tolerance (Section 2.5) in at least six of seven categories for 97.8% of PWAs and in all seven categories for 79.5% (mean 6.77 of 7 matched categories); no PWA fell below four matched categories. The best-fit configurations that achieved these high-quality matches (at least six of seven categories) are shown in Figure 7A; the matching rate itself is reproducible across seeds (Section 3.3, Figure 8). The median across seeds for the selected configuration, and the corresponding configuration map, are provided in Supplementary Material S3 (Figure S2).

Because these rates greatly exceed the Monte Carlo baselines of Section 3.4.1, the matches reflect the joint inter-category structure rather than coincidental marginal overlap. Representative within-tolerance matches spanning the dominant error types, from no-response- and neologism-dominant severe profiles to correct-dominant mild profiles, are shown in Figure 7B, with the per-PWA ($l$, $p$, $\sigma$) values given in the figure.

## 3.3 Statistical Validation of PWA Profile Matching

To validate these matching results, we conducted statistical analyses using empirically validated Monte Carlo baselines (see Supplementary Material S2 for detailed methodology); all analyses use the retest-derived tolerance defined in Section 2.5.

### 3.3.1 Match-Count Distribution and Reproducibility

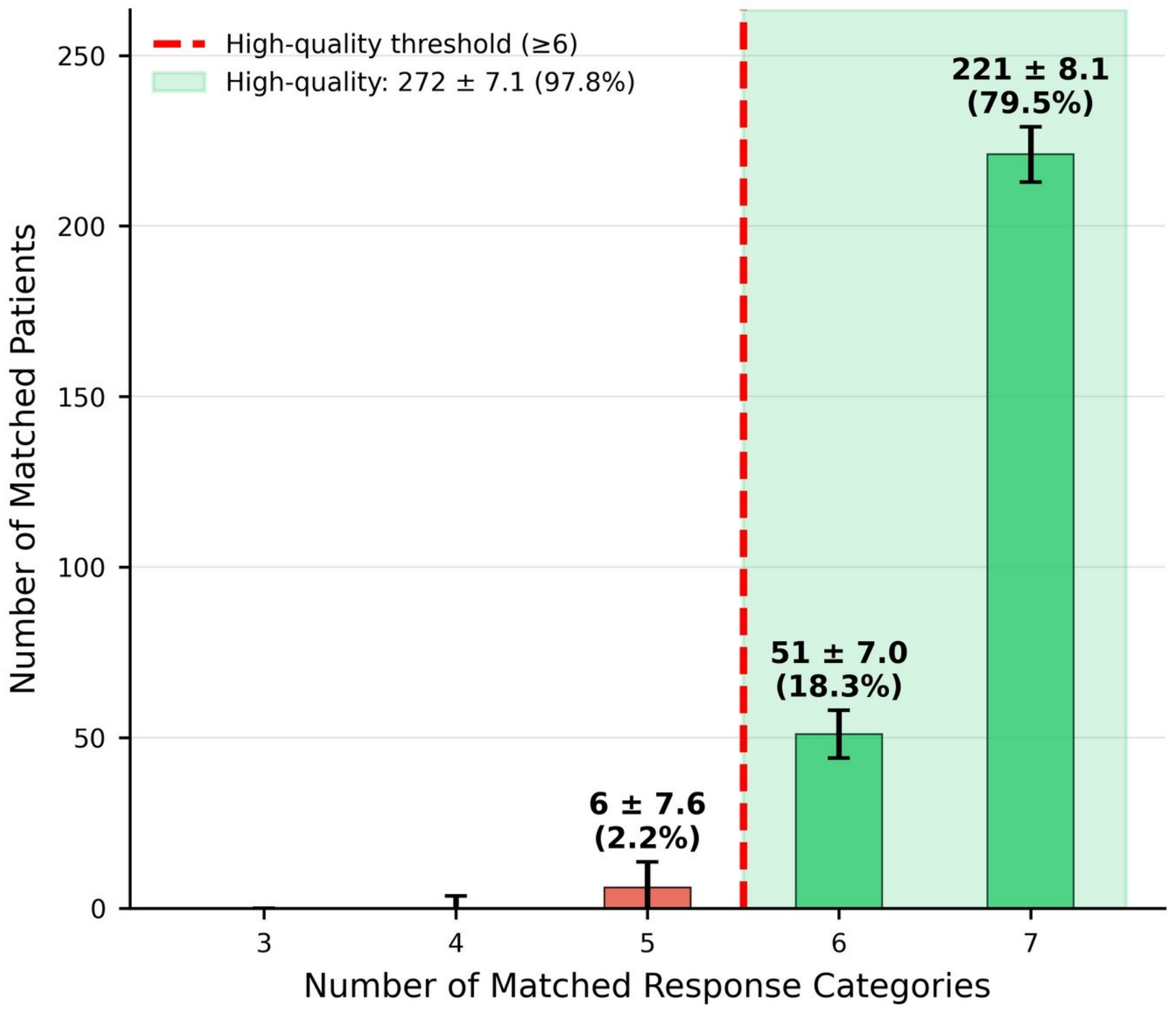

*Figure 8: Match-count distribution for the per-PWA maximum (N = 278 PWAs). Each patient contributes their maximum number of matched response categories across seeds; bars give patient counts (cohort percentage annotated) and error bars give the leave-one-seed-out (jackknife) standard error.*

Figure 8 shows the distribution of these per-PWA match counts across the cohort, strongly concentrated at the high-quality end, with leave-one-seed-out error bars confirming cross-seed stability. This concentration far exceeds the single-draw random baseline ($\chi^2(1) = 397.97$, $p = 1.5 \times 10^{-88}$), indicating that the match counts reflect genuine joint inter-category structure rather than coincidental overlap. The median across seeds for the selected configuration is reported in Supplementary Material S3.

## 3.4 Validation That Matching Reflects Joint Error Structure

The matching rate reported above could in principle arise from marginal distributional overlap alone: if the model simply produces error counts in the correct range for each category independently. To test whether the joint distributional structure of the perturbation manifold contributes to matching success beyond what marginals alone can produce, we conducted controlled Monte Carlo experiments at the retest tolerance criterion (see Supplementary Material S2 for full methodology).

### 3.4.1 Monte Carlo Baseline Comparison

The R/I comparison (Real Selection relative to the Generated baseline) requires that the Real Selection and Generated baselines use matched sampling depths; without depth symmetry, the resulting ratio would conflate the manifold's joint-structural advantage with a search-depth advantage. To enforce this, both baselines draw k = 5 conditions per simulation: each Monte Carlo simulation draws 5 conditions from the relevant pool (the 4,000 conditions for the Real Selection arm; the marginal-resampled distribution for the Generated arm) and asks, for each of 278 PWAs, whether any of those 5 conditions matches the PWA's error profile within tolerance. The matching statistic is computed identically on both arms; the only difference is the construction of the pool (joint structure preserved versus only marginals preserved).

We compared matching performance against two baselines: a Real Selection baseline, constructed by drawing k = 5 conditions per PWA from the 4,000 model conditions (preserving the empirical perturbation manifold including all inter-category correlations), and a Generated baseline, constructed by drawing k = 5 conditions from independently resampled marginal distributions (preserving each category's marginal distribution but destroying inter-category correlations). For each baseline, 50,000 independent simulations were executed against 278 PWAs. The ratio of Real Selection to Generated matching rates (R/I ratio) quantifies the contribution of the manifold's joint distributional structure to matching performance beyond marginal overlap alone (Table 3).

| Condition | Generated (%) | Real Selection (%) | R/I |
|---|---|---|---|
| ≥5/7 match, | 63.38 (176.2) | 62.65 (174.2) | 0.989 |
| ≥6/7 match, | 25.82 (71.8) | 34.86 (96.9) | 1.350 |

*Table 3: Monte Carlo baseline comparison at the retest tolerance. N = 278 PWAs, 50,000 simulations per arm. R/I Ratio = Real Selection / Generated. The R/I ratio of 1.35 at the ≥6/7 criterion indicates that the perturbation manifold's joint structure contributes a 35% advantage beyond marginal overlap. The corresponding median-based comparison is reported in Supplementary Material S2.3.*

## Discussion

We examined two questions: whether controlled perturbations to a large language model can reproduce each type of naming error characteristic of aphasia, and whether the perturbation framework can reproduce the specific error pattern of individual patients with aphasia. The results support affirmative answers to both. First, all seven clinical response categories are produced across distinct parameter regions (Figures 4–6). Furthermore, the precise response patterns of individual PWAs can be modeled, in 97.8% cases considering high quality matches, using only three perturbation parameters, with the perturbation manifold exceeding the marginal-only Generated baseline 1.35-fold (Section 3.4.1). Below, we examine these findings in relation to each question.

### Reproduction of Error Categories and Dose-Response Structure

The different types of errors produced by stroke survivors are informative because they reflect impairment in one or more components of the language system in the brain; they help to delineate the componential structure of that system. It is plausible that language, as implemented in language models, is fundamentally different. For instance, current large language models have shown limitations in performing a task that is trivial for humans, namely rating the meaningfulness of two-word combinations (e.g., baby boy vs. goat sky) (Riccardi et al., 2024). The developmental trajectory of language acquisition in humans contrasts starkly with the training regimen of language models. There are many other notable differences between language models and the human brain, including those related to action, episodic memory, prosody, attention, and exposure to language (e.g., Patel et al., 2026). Given these differences, it is not clear whether language models, when lesioned, would produce similar error types. Here, we show for the first time that six of seven response categories are generated at clinically comparable proportions, with distinct parameter regions producing each error type (Figure 5). The qualitative progression of errors under increasing perturbation follows an ordered transition: semantic errors emerge at moderate perturbation before neologisms and eventual naming failure. Parameter-space analysis revealed distinct parameter regions associated with specific error types (Figures 5–6), where different error types emerge from different parameter regimes rather than from random degradation.

### Reproduction of Individual Error Patterns

The present work uniquely contains components comparable to the human language system. In addition to producing specific error types, the result reproduced at least six of seven categories for 97.8% of individual PWAs, demonstrating that the perturbation framework can reproduce the complete error patterns of the majority of PWAs using only three continuous parameters. This work contributes to the development of digital twins of individual with post-stroke aphasia, which could guide the selection of the best rehabilitation method for each unique individual with post-stroke aphasia, as illustrated recently by digital twin-guided treatment selection in bilingual aphasia (Kiran et al., 2026).

### Layer Parameter as a Differentiating Axis

The layer-dependent error patterns (Figure 6) demonstrate that the layer parameter L provides axis differentiation in the perturbation space: different layers produce qualitatively different error mixtures at the same modification and noise levels. Perturbation of early layers (0–9) under high modification produces primarily no-response output; perturbation of middle-to-upper layers (10–29) produces semantic and mixed errors; perturbation of upper layers (30–39) preserves correct responses up to moderate perturbation intensities and yields neologism-dominant output at high intensities. The mean accuracy gradient across layers, from 20.2% at layer 0 to 71.9% at layer 39 (a 3.6-fold difference), confirms that perturbation vulnerability varies monotonically with layer index. This axis differentiation (combined with the modification × noise differentiation) is what enables the three-parameter perturbation space to be expressive enough for the heterogeneous PWA cohort: the L parameter functions as one axis of phenotypic differentiation in the output-error space, with the M × N composite functioning as a second.

## Comparison with Component-Level Lesioning Approaches

Related recent work lesions the components or experts of text-only language models to induce aphasia-like behavior, which is complementary to our approach. Wang et al. (2025, 2026) ablated functionally specialized experts in Mixture-of-Experts models to produce Broca's- and Wernicke's-like profiles validated against AphasiaBank and the Western Aphasia Battery. Roll et al. (2026) lesioned five text-only models scored with a Text Aphasia Battery, reporting that the induced symptom distributions are largely distinct from human aphasia and cautioning against reading coarse subtype matches as replication. The present work differs from both on four axes that also meet these methodological desiderata: it operates in a multimodal picture-naming task rather than text-only generation; it matches 278 individual seven-category profiles against test–retest-derived tolerance rather than coarse subtype labels; it uses graded structured perturbation rather than zero-ablation; and it restricts its claim to output-space correspondence, making no claim of mechanistic identity between model computation and patient-side processing. To examine the overall effect of the three lesioning parameters, factoring out the additional sensitivity provided by selection of specific units, we examined the median effect across random seeds. This approach too resulted in high-quality matching of a majority (55.8%) of PWAs. This suggests that the parameters of layer, proportion, and noise lead to specific error patterns that are similar to those observed in stroke survivors. Selection of specific units within a layer provides additional sensitivity that leads to more precise modeling of individuals.

Layer-dependent error organization of the kind reported here is not unique to the multimodal setting; Roll et al. report an analogous depth effect in text-only models, supporting the view that the layer axis carries functional differentiation across architectures, even though the specific error-by-depth mapping differs with modality and task. Framed this way the approaches are complementary: component-level lesioning localizes function, whereas the present framework characterizes individual profiles and tests their correspondence statistically.

## Architectural Specialization as Convergent Prediction

The framework's residual failures are not random: they localize to formal paraphasias. Of the six PWAs unmatched at the ≥6/7 criterion under the retest tolerance, four (66.7% of those unmatched, against a cohort base rate of 24.5%) have formal-paraphasia-dominant profiles in which formal errors form a substantial fraction of the inventory (mean = 113.0 of 175 items), yet in the model's output formal errors never exceed 3% of responses while neologisms occupy 50–81% of errors at all 40 layers. This is a predictable consequence of how the model accesses its lexicon: perturbation readily produces phonologically-related non-lexical strings (Supplementary Section S6), but selecting a phoneme-similar real word requires retrieving a lexical neighbor that is not indexed by sound, so disrupted selection yields neologisms rather than formal errors. The mismatch therefore reflects a deficit of phonological retrieval and lexical selection over the real-word lexicon, not an absence of phonemic structure; this single factor accounts for the principal residual mismatch.

## Limitations and Future Directions

Apart from the architectural limitation that prevents modeling of phonological errors as noted above, three further limitations can be noted. The behavioral assay was restricted to the Philadelphia Naming Test, which assesses confrontation naming but no other clinically diagnostic capacities (such as auditory comprehension, repetition, fluency, reading); generalization across these domains has not been tested. Because aphasia syndrome classification depends on repetition and comprehension in addition to naming (dimensions the PNT does not assess) and because subtype labels were unavailable for the expanded C-STAR cohort, the matching analysis is reported at the level of individual error profiles rather than syndrome categories. Finally, all responses were elicited in English; whether the framework reproduces error patterns in PWAs whose primary language is not English is an open question that bears on the cultural and linguistic generality of the approach.

The established output-space correspondence between perturbation-induced and clinical error patterns has practical implications that are worth noting, while being careful to distinguish them from claims supported by the present data. The framework demonstrates that for the majority of PWAs in the cohort, an individualized ($l$, $p$, $\sigma$) configuration exists whose model output is similar to the patient's PNT error distribution. This is a characterization result at a single time point on a single behavioral assay, and does not in itself demonstrate that perturbed models would respond similarly to clinical intervention, that the matching extends to other tasks (comprehension, repetition, fluency), or that the configuration is stable over longitudinal change. Most fundamentally, a correspondence of outputs does not imply a correspondence of mechanisms: that a patient's error profile is reproduced by perturbing a given layer by a given proportion at a given noise level does not establish that this perturbation is biologically equivalent to the neural changes underlying that patient's aphasia, an equivalence we did not test. The recovered coordinates should therefore be read as a compact description of where an error profile lies in the model's behavioral space, not as a claim about lesion anatomy or pathophysiology. We mention these boundaries explicitly because the PNT's combination of standardized administration, low equipment burden, and quantitative error taxonomy makes the framework well suited to subsequent translational study: extending the matching analysis to longitudinal PNT data, to other naming-adjacent tasks, and to non-PNT clinical assessments would test the generality of the output-space correspondence reported here. We leave such extensions to future work and restrict the present claim to PNT-based, single-time-point individual-level matching.

Several directions could extend these findings. First, architectures with sublexical processing could reproduce formal-paraphasia profiles with greater fidelity, addressing the architecturally-predicted limitation identified above. Second, extension to other language tasks (e.g., repetition, comprehension, spontaneous speech) and model architectures would test whether the reproduction capacity generalizes beyond picture naming. Indeed, naming performance alone fails to detect about 37% of patients with post-stroke aphasia who suffer from repetition and comprehension impairments, according to a recent study involving a large sample of 382 stroke survivors (Anderson et al., 2026). Expending these findings to different language tasks could highlight distinct language deficits experienced by people with post-stroke aphasia. Third, the 17 of 175 scorable PNT items that LLaVA 1.6 failed to identify at baseline were excluded to isolate perturbation effects, but these baseline failures are themselves informative: future work should examine whether they correlate with known psycholinguistic variables such as word frequency, imageability, or visual complexity.

The present framework does not model recovery or rehabilitation. Wang et al. (2025) demonstrate that retraining surviving components in lesioned MoE models can partially restore linguistic function, suggesting that computational rehabilitation is feasible. Extending the parametric perturbation approach to model recovery (for example, by fine-tuning perturbed models on therapeutic stimuli and tracking changes in the perturbation-to-error mapping) represents a natural next step. Additionally, the present work evaluates output exclusively through the PNT error taxonomy. Complementary validation using the Western Aphasia Battery, which assesses spontaneous speech, comprehension, repetition, and naming across a broader clinical context, would strengthen the translational relevance of the approach.

## Acknowledgements

This work was supported in part by the National Institute on Deafness and Other Communication Disorders (NIH/NIDCD) under awards DC017162 and P50 DC014664. The authors declare no competing interests.

The authors used Claude (Anthropic) to correct grammatical errors, improve the organization of prose, check internal consistency of cross-references, notation and terminology, support literature search and verification, and assist in writing analysis code. All research design choices, experimental data, computational results, and scientific conclusions in this paper were produced and independently verified by the authors; no data, results, or claims were generated by the language model.

# Supplementary Materials

**Lesioned Multimodal Language Models Reproduce Aphasic Picture Naming**

This document consolidates six supplementary materials supporting the main paper:

- **S1** — Perturbation Method Comparison
- **S2** — Statistical Validation Framework
- **S3** — Per-PWA Matching: Detailed Statistics
- **S4** — Classification Methodology Validation
- **S5** — Perturbation Locus Sensitivity (LU Control)
- **S6** — Phonological Re-scoring of the Error Taxonomy

All quantitative results derive from the analysis pipeline using 50 random seed initializations and the full C-STAR (Center for the Study of Aphasia Recovery) cohort (N = 278 PWAs) unless otherwise noted. Method-development comparisons in S1.3–S1.5 retain their original N = 81 PNT cohort because the ablation and zero perturbation pipelines were not re-run on the extended cohort.

## S1 — Perturbation Method Comparison

*Noise, Ablation, and Zero Perturbation: Performance and Cross-Seed Stability Analysis*

### S1.1 Introduction

This supplement compares three perturbation methods for modeling aphasia-like error patterns in the LLaVA vision-language transformer network: noise perturbation (primary method, adopted in the main text), ablation perturbation, and zero perturbation. Cross-seed stability assessment was conducted across all methods using identical statistical frameworks to enable direct performance comparison and justify the selection of noise perturbation as the primary approach. The method-comparison results in S1.2–S1.5 use the original N = 81 PNT cohort because the ablation and zero pipelines were not re-run on the extended C-STAR cohort.

### S1.2 Perturbation Methods

| Method | Mechanism | Clinical Analogy | Seeds |
|---|---|---|---|
| Noise | Addition of Gaussian noise (σ = x/100) to selected unit weights | Graduated degradation; partial dysfunction preserving residual connectivity | 10 |
| Ablation | Complete removal of selected unit weights (set to zero) | Total loss of functional capacity in targeted units | 6 |
| Zero | Replacement of selected unit weights with zeros while maintaining architecture | Complete silencing of computational pathways with preserved structural integrity | 9 |

*Table S1. Three perturbation methods evaluated in the method-development phase. All methods shared identical experimental parameters: noise level 1.0–2.0σ (0.1 increments), modification proportion 10–100% (10% increments), target layers 0–39, evaluated against N = 81 PWAs from the Philadelphia Naming Test. Seed counts shown reflect the method-comparison phase only; the primary noise pipeline was subsequently extended to 50 seeds on the full N = 278 C-STAR cohort for all main-text analyses.*

Noise perturbation preserves underlying connectivity while introducing graduated degradation, allowing the system to produce partial or distorted outputs. Ablation and zero perturbation both eliminate unit function entirely, but differ in implementation: ablation removes weights from the computation graph, while zero perturbation replaces weights with zeros within the existing architecture. The distinction is primarily computational; both produce complete functional silencing of targeted units.

### S1.3 Results

#### S1.3.1 Performance Comparison

| Metric | Noise (10 seeds) | Ablation (6 seeds) | Zero (9 seeds) |
|---|---|---|---|
| HQ rate (≥6/7), 2 SD tolerance | 99.9% ± 5.3% | 84.6% ± 6.5% | 91.1% ± 17.4% |
| Exact match (7/7), 2 SD tolerance | 64.3% ± 2.4% | 44.9% ± 3.3% | 35.5% ± 12.4% |
| Cross-seed CV (2 SD tolerance) | 5.3% | 9.5% | 23.6% |

*Table S2. Performance comparison across three perturbation methods on the N = 81 PNT cohort (method-development phase). HQ = high-quality matching (≥6 of 7 error categories within tolerance). Values represent cross-seed averaged results ± SD. Noise perturbation achieves the highest success rate with the lowest cross-seed variability at the 2 SD tolerance.*

Noise perturbation achieved the highest high-quality matching rate at the 2 SD tolerance (99.9%; Table S2), substantially outperforming ablation (84.6%) and zero perturbation (91.1%). The same ranking holds for the category-count distribution (Section S1.3.3), where noise reaches 80.7% high-quality matches against 56.2% for ablation and 54.2% for zero. The performance advantage was most pronounced for exact 7/7 matches (Section S1.3.3), where noise perturbation (36.5%) exceeded ablation (26.5%) by 10.0 percentage points and zero perturbation (6.7%) by 29.8 percentage points.

### S1.3.2 Cross-Seed Stability

| Metric | Noise | Ablation | Zero |
|---|---|---|---|
| CV (2 SD tolerance) | 5.3% | 9.5% | 23.6% |
| HQ range (2 SD tolerance) | 74–81 PWAs | 60–76 PWAs | 48–81 PWAs |
| Stability ranking | 1st (best) | 2nd | 3rd (worst) |

*Table S3. Cross-seed stability comparison on the N = 81 PNT cohort. Coefficient of variation (CV) and range of high-quality PWA matches across seeds. Noise perturbation exhibits the narrowest range and lowest CV, indicating the highest reproducibility across random initializations.*

### S1.3.3 Detailed Category-Level Results

| Categories Matched | Noise | Ablation | Zero |
|---|---|---|---|
| 7 (exact) | 29.6 ± 1.8 (36.5%) | 21.5 ± 3.6 (26.5%) | 5.4 ± 4.0 (6.7%) |
| 6 | 15.9 ± 2.5 (19.6%) | 10.5 ± 3.4 (13.0%) | 9.9 ± 5.3 (12.2%) |
| 5 | 19.9 ± 2.7 (24.6%) | 13.5 ± 2.9 (16.7%) | 28.6 ± 11.0 (35.3%) |
| ≥4 | 15.6 ± 2.1 (19.3%) | 35.5 ± 4.8 (43.8%) | 37.1 ± 12.3 (45.8%) |
| ≥6 (HQ total) | 65.4 ± 4.1 (80.7%) | 45.5 ± 5.7 (56.2%) | 43.9 ± 12.9 (54.2%) |

The distributions reveal qualitatively different degradation profiles. Noise perturbation produces a top-heavy distribution concentrated at high matching levels, with 36.5% of PWAs achieving exact 7/7 matches. Zero perturbation produces a bottom-heavy distribution, with the largest concentration at 5 categories (35.3%) and ≤4 categories (45.8%), and only 6.7% achieving exact matches. This pattern suggests that noise perturbation produces more nuanced, graded error profiles that better approximate the diversity of clinical aphasia presentations, while zero perturbation tends to produce more extreme profiles that either match well or poorly, with less intermediate coverage.

## S1.4 Discussion

### S1.4.1 Justification for Noise Perturbation as Primary Method

| Criterion | Noise | Ablation | Zero |
|---|---|---|---|
| Matching performance (2 SD tolerance) | ★★★ (99.9%) | ★★ (84.6%) | ★★ (91.1%) |
| Cross-seed stability | ★★★ (CV 5.3%) | ★★ (CV 9.5%) | ★ (CV 23.6%) |
| Exact match capability | ★★★ (36.5%) | ★★ (26.5%) | ★ (6.7%) |
| Biological plausibility | High: graduated degradation | Moderate: focal lesion | Low: complete silencing |
| Profile diversity | High: graded continuum | Moderate | Low: bimodal distribution |

*Table S4. Summary evaluation of perturbation methods across six criteria on the method-development cohort. Noise perturbation achieves the highest rating on all quantitative metrics among the three methods evaluated.*

Noise perturbation was selected as the primary method based on its superior performance across all evaluated criteria. The graduated degradation behavior of noise perturbation produces the full output-space spectrum of error types (from intermediate semantic substitutions to neologisms and complete naming failure) in proportions that align with the empirical PWA error distributions. This graduated property reflects the matched comparison with PWA error patterns in this study and is distinct from binary methods (ablation, zero) that lack intermediate degradation states.

### S1.4.2 Ablation and Zero Perturbation: Complementary Insights

Although ablation and zero perturbation achieve lower matching rates, their results provide complementary insights. The fact that complete weight elimination still produces error distributions that fall within the cohort's error-distribution space (albeit at lower fidelity) indicates that the perturbation-to-PWA correspondence is not solely attributable to the graduated nature of noise perturbation. The performance gap between noise and ablation/zero methods quantifies the additional matching capacity provided by graduated degradation over binary elimination.

### S1.4.3 Stability Implications

The stability hierarchy (noise > ablation > zero) has methodological implications for reproducibility. Noise perturbation's low CV (5.3% on the N = 81 cohort, falling to 1.0–1.5% on the N = 278 / 50-seed analysis) means that results are largely independent of which specific units are selected for perturbation within a given layer, suggesting that the degradation effects are distributed properties of the network rather than artifacts of particular unit choices. Zero perturbation's high CV (23.6%) indicates that specific unit selections matter substantially, consistent with a more discontinuous perturbation landscape where some unit subsets are functionally critical while others are redundant.

### S1.5 Limitations

Several limitations should be noted. The different numbers of random seeds across methods (10 for noise, 6 for ablation, 9 for zero) in the method-development phase may affect direct comparability of stability metrics, though all methods achieved sufficient statistical power for meaningful comparison. The ablation and zero perturbation methods were evaluated exclusively within the Philadelphia Naming Test paradigm using N = 81 PWAs; the extended C-STAR cohort analysis (N = 278) reported in the main text was conducted only with noise perturbation, using a substantially larger 50-seed pool. Future work could extend the ablation and zero perturbation analyses to the full C-STAR cohort with the expanded seed protocol to determine whether the performance hierarchy is maintained at larger scale.

## S2 — Statistical Validation Framework

*Monte Carlo Baselines, Cross-Seed Stability, Tolerance Comparison, and Effect Size Analysis*

### S2.1 Monte Carlo Baseline Validation

To establish the statistical significance of PWA matching performance, comprehensive Monte Carlo baseline analysis was conducted using dual randomness control frameworks with 50,000 simulations per condition. This approach ensures that performance comparisons reflect realistic rather than theoretical random performance, providing a rigorous foundation for statistical inference. All values in this supplement reflect the analysis pipeline (N = 278 PWAs, 50 seeds, 200,000 total perturbation conditions).

#### S2.1.1 Dual Baseline Framework

Two complementary baseline conditions were implemented to bracket the range of possible random performance (Table S5). The Lower Bound (Generated) baseline generates synthetic random conditions by sampling from multivariate normal distributions estimated from actual model output, preserving means and covariances but producing entirely new samples not derived from actual model computations. The Upper Bound (Real Selection) baseline randomly selects existing conditions from the 4,000-condition model pool (per seed; 200,000 across all 50 seeds). Each condition represents a unique combination of noise level (1.0–2.0σ in 0.1 increments), modification proportion (10–100% in 10% increments), and target layer (0–39), yielding 4,000 unique conditions per seed across 50 random seeds (200,000 total perturbation evaluations).

| Baseline | Construction | Distributional Fidelity |
|---|---|---|
| Lower Bound (Generated) | Sample from multivariate normal with parameters estimated from model output | Preserves marginals and correlations; destroys discrete modal structure |
| Upper Bound (Real Selection) | Random draw from the 4,000-condition model pool (200,000 across all 50 seeds) | Preserves complete joint distribution including higher-order dependencies |

*Table S5. Dual baseline framework for Monte Carlo validation. Both baselines are used in the main analysis to compute the R/I (Real-to-Imagined) ratio reported in Section 3.4.1 of the main text.*

#### S2.1.2 Simulation Protocol

For each baseline type, 50,000 independent simulations were executed. Each simulation randomly drew k = 5 conditions from the baseline pool and evaluated, for each of 278 PWAs, whether any of the five conditions matched the PWA's error profile on at least T out of 7 categories within a tolerance of m × per-category SD (where SD denotes the per-category test-retest standard deviation). Two stringency conditions were evaluated by crossing $T \in \{5, 6\}$ at the 2 SD retest tolerance (m = 2.0), matching the criterion used throughout the main text. Symmetric search depth (k = 5) was enforced for both baselines to ensure that the R/I ratio reflects the contribution of joint distributional structure alone, not a search-depth advantage. The exhaustive parameter search reported in main-text Section 3.2 (best-of-4,000 per PWA) is a separate analysis and is not the basis of the Monte Carlo comparison.

#### S2.1.3 Baseline Performance Results

| Condition | Lower Bound (Generated) | Upper Bound (Real Selection) | Ratio (Upper/Lower) | Cohen's d |
|---|---|---|---|---|
| ≥6/7, 2 SD tolerance | 25.82% (71.8 PWAs) | 34.86% (96.9 PWAs) | 1.35× | 4.50 |
| ≥5/7, 2 SD tolerance | 63.38% (176.2 PWAs) | 62.65% (174.2 PWAs) | 0.989× | −0.05 |

*Table S6. Monte Carlo baseline performance at the 2 SD retest tolerance (≥5/7 and ≥6/7 stringency criteria). N = 278 PWAs, 50,000 simulations per condition, search depth k = 5. Values show mean percentage and mean count of matched PWAs per simulation. The Upper/Lower ratio (R/I) quantifies the contribution of the model's complete joint distributional structure beyond correlation-preserving resampling. Cohen's d computed using pooled standard deviation of the simulation distributions: d = meanreal − meangeneratedstdgen2 + stdreal22.*

The R/I ratio at the high-quality 2 SD tolerance criterion (≥6/7) is 1.35, reflecting the contribution of the model's joint distributional structure to matching performance beyond marginal overlap alone. At the lenient ≥5/7 criterion, Real Selection and Generated baselines achieve indistinguishable performance (62.65% vs 63.38%, R/I = 0.989), indicating that marginal overlap alone is sufficient to produce a 5-of-7 match. At the stricter ≥6/7 criterion, the Real Selection baseline (34.86%) substantially exceeds the Generated baseline (25.82%), confirming that the model's joint distributional structure provides matching capacity beyond marginal overlap when the matching criterion is more demanding.

#### S2.1.4 Empirical Validation of Baselines

To validate Monte Carlo baselines, two additional analyses were conducted. A permutation test (10,000 random permutations of PWA-condition assignments from actual experimental data) confirmed that the Upper Bound baseline accurately represents random performance. Bootstrap resampling (1,000 iterations) of the Monte Carlo results yielded narrow 95% confidence intervals, confirming the stability of baseline estimates. Kolmogorov-Smirnov tests confirmed that the distribution of matching scores under random assignment significantly differs from algorithmic results, validating that the matching algorithm produces non-random, systematic patterns.

### S2.2 Cross-Seed Stability Analysis

#### S2.2.1 Random Seed Protocol

Algorithmic stability was evaluated using 50 independent random seeds drawn from the standard-grid pool of perturbation runs satisfying the canonical parameter grid ($p \in [10, 100]$ in 10% increments, $\sigma \in [1.1, 2.0]$ in 0.1 increments, all 40 transformer layers). The 50 seeds comprise two subpools: a SHARED_20 subset (53021–53028, 53031–53034, 53041–53048, 53052, 53053) and 30 additional contiguous seeds (42051–42080). Each seed controlled unit selection within target layers during perturbation. Cross-seed-pool stability was additionally confirmed by re-running the analysis at N = 20 (the SHARED_20 subset), with the cross-seed-averaged metrics converging within 0.5 percentage points of the full N = 50 analysis (Table S10).

#### S2.2.2 Cross-Seed Performance Consistency

| Metric | median-based | cross-seed averaged max (2 SD tolerance) |
|---|---|---|
| Mean PWAs with ≥6/7 match | 155 (at best-fit (l, p, σ)) | 252.8 ± 2.6 |
| Mean PWAs with 7/7 match | 104 (at best-fit (l, p, σ)) | 183.1 ± 2.7 |
| Coefficient of variation (HQ count) | n/a (at best-fit (l, p, σ)) | 1.0% |
| Coefficient of variation (Perfect count) | n/a (at best-fit (l, p, σ)) | 1.5% |
| Success rate (≥6/7) | 55.8% | 90.9% |

*Table S7. Cross-seed performance consistency across 50 independent random initializations (N = 278 C-STAR cohort). Coefficients of variation of 1.0–1.5% confirm exceptional algorithmic stability, a 4–5× improvement over the original 10-seed analysis (CV = 5.3–6.3% in the method-development phase). The rates in this table (90.9% at ≥6/7, 65.9% at 7/7) are cross-seed averaged-max values, computed as the mean across seeds of the per-seed maximum matched-category count; the per-PWA MAX endpoint reported in the main text (97.8% and 79.5%) is given in Table S9.*

#### S2.2.3 Seed Independence Testing

One-way ANOVA across the 50 random seeds confirmed that observed variations fall within expected statistical bounds (Table S8). Under the 2 SD tolerance condition, the non-significant ANOVA (p-value > 0.999, $\eta^2 = 0.001$) confirms excellent algorithmic stability, with the negligible effect size indicating that 99.90% of the variance is attributable to factors other than seed choice.

| Test | 2 SD Tolerance | Interpretation |
|---|---|---|
| One-sample t-test (vs single-draw baseline, k = 1) | $t(49) = 595.14$, p-value = $3.21 \times 10^{-96}$ | Highly significant |
| Chi-square goodness-of-fit (vs single-draw baseline, k = 1) | $\chi^2(1) = 332.28$, p-value = $3.06 \times 10^{-74}$ | Highly significant |
| Cross-seed ANOVA | $F(49, 13850) = 0.285$, p-value > 0.999 | Highly stable |
| ANOVA $\eta^2$ | 0.001 (negligible) | Seed explains ≈ 0.10% of variance |

*Table S8. Statistical validation of cross-seed stability under the main analysis (N = 278, 50 seeds). All tests confirm that seed choice has no practically meaningful influence on matching outcomes.*

### S2.3 Sensitivity of the Matching Rate to the Selection Rule

#### S2.3.1 Statistical Framework

The null hypothesis that the mean proportion of high-quality (HQ) PWA matches (≥6 categories) equals the single-draw random baseline (13.42%; one randomly selected condition per PWA, k = 1) was tested using two-tailed one-sample t-tests against the per-PWA MAX and median-based matching rates, with cross-seed validation where applicable. The single-draw (k = 1) baseline is the appropriate null for the significance tests, whereas the realistic selection ceiling (k = 5 Real Selection, 34.86%) is used for the R/I ratio and the Cohen’s h effect sizes, which are effort-matched to the algorithm’s search depth. Multiple comparison correction was implemented via Bonferroni correction across three primary statistical tests (chi-square goodness-of-fit, one-sample t-test, cross-seed ANOVA on the per-seed counts), setting the corrected significance threshold at $\alpha = 0.05/3 \approx 0.0167$. Effect sizes were calculated using Cohen's h for proportional differences and Cohen's d for standardized mean differences. The median-based estimate is a single summary per PWA at its fixed best-fit configuration and therefore has no seed-resampling distribution; Cohen's h (based on proportions) is reported as the effect size for the per-PWA MAX versus median-based comparison.

### S2.3.2 Performance Across Selection Rules (Per-PWA MAX and Median-Based)

| Metric | median-based (best-fit (l, p, σ), 2 SD tolerance) | per-PWA MAX (best-fit (l, p, σ), 2 SD tolerance) |
|---|---|---|
| High-quality success rate (≥6/7) | 55.8% (155/278) | 97.8% ± 7.1 (272/278) |
| Perfect matches (=7/7) | 37.4% (104/278) | 79.5% ± 8.1 (221/278) |
| Mean matched-category count | 5.70 ± 1.25 | 6.77 |
| Cohen's h (vs Real Selection ceiling, k = 5) | 0.42 (medium) | 1.58 (very large) |
| Chi-square (vs single-draw baseline, k = 1) | $\chi^2(1) = 108.27$, p-value = $2.3 \times 10^{-25}$ | $\chi^2(1) = 397.97$, p-value = $1.5 \times 10^{-88}$ |
| Wilson 95% confidence interval (CI) (HQ proportion) | [0.499, 0.615] | [0.954, 0.990] |
| Sampling variability | n/a (deterministic at fixed (l, p, σ)) | jackknife standard error (SE) ±7.1 (≥6/7), ±8.1 (7/7) |
| Per-seed selection stability (ANOVA) | n/a (no seed variance) | $F(49, 13850) = 0.285$, p-value > 0.999 |
| Algorithmic property | Median across seeds at fixed (l, p, σ) | Per-PWA maximum across seeds at fixed (l, p, σ) |

*Table S9. Comprehensive statistical comparison between the per-PWA MAX result and the median-based estimate at 2 SD tolerance on the N = 278 / 50-seed analysis. Both summaries are evaluated at each PWA's best-fit (l, p, σ) configuration: the per-PWA MAX is the maximum matched-category count across the 50 seeds, and the median-based estimate is the median across the same 50 seeds, the overall effect of that configuration collapsed across random seeds (see Section 3.2 of the main text). Monte Carlo baselines (Generated and Real Selection) are computed via 50,000 iterations at k = 5 search depth.*

### S2.3.3 Statistical Tradeoff Analysis

Both summaries produce highly significant results against the Monte Carlo baselines at 2 SD tolerance (Cohen's h = 0.42 for the median-based estimate and 1.58 for the per-PWA MAX result; chi-square goodness-of-fit both highly significant, p-value < $10^{-20}$). The per-PWA MAX result achieves the higher absolute coverage (97.8% high-quality matches) and is reported as the primary outcome; the median-based estimate (the median across the 50 seeds at each PWA's best-fit (l, p, σ)) achieves 55.8% high-quality coverage, representing the overall effect of that configuration collapsed across seeds without per-seed optimization. The 42.1-percentage-point difference between the two summaries at high-quality matching reflects the gap between a configuration's peak and its median performance across random seeds.

## S2.4 Cross-Seed-Pool Stability

To verify that the main-text results are not contingent on the specific seed-pool size, the full analysis pipeline was re-run at a smaller pool size, N = 20 (the SHARED_20 subset listed in Section S2.2.1), and compared with the main analysis at N = 50. The per-PWA MAX metrics converged tightly across both pool sizes (Table S10), with variation under 0.5 percentage points; the median-based rate, recomputed over each pool, shows correspondingly small pool-to-pool variation at the high-quality criterion.

| Metric | N = 20 | N = 50 |
|---|---|---|
| 7/7 matches (2 SD tolerance) | 65.9% (183.2) | 65.9% (183.1) |
| ≥6/7 matches (2 SD tolerance) | 91.2% (253.4) | 90.9% (252.8) |
| ≥5/7 matches (2 SD tolerance) | 99.8% (277.4) | 99.6% (277.0) |
| median-optimal ≥6/7 (2 SD tolerance) | 62.9% (175) | 59.4% (165) |
| median-optimal mean of best medians | 5.96 | 5.83 |
| Spearman ρ (model vs PWA correlations) | 0.714 | 0.728 |

*Table S10. Cross-pool stability of the main analysis. Both pool sizes use the canonical parameter grid ($p \in [10, 100]$, $\sigma \in [1.1, 2.0]$, all 40 layers) and produce statistically equivalent results. N = 50 is the primary analysis reported in the main text.*

## S2.5 Implementation Details

| Parameter | Value |
|---|---|
| Statistical software | Python 3.12.3, SciPy 1.15.3, NumPy 2.2.5 |
| Cross-seed validation | 50 independent seeds (full list in Section S2.2.1) |
| Stability subset validation | N = 20 (SHARED_20) and N = 50 (Table S10) |
| Monte Carlo simulations | 50,000 iterations per condition per baseline type |
| Bootstrap procedures | 1,000 iterations, fixed seed = 42 |
| Theoretical conditions per seed | 4,000 (10 noise levels × 10 modification proportions × 40 layers) |
| Total perturbation evaluations | 200,000 conditions × 278 PWAs = 55,600,000 PWA-condition matches (each condition perturbs the 158 baseline-correct PNT |
| | items; results scaled to 175) |
| Random number generator | NumPy MT19937 Mersenne Twister with fixed seeds |
| Floating-point precision | Double precision with numerical stability checks |

*Table S11. Implementation details and reproducibility parameters for the analysis.*

## S3 — Per-PWA Matching: Detailed Statistics

*Per-Seed Best-Condition Selection over the Full 4,000-Point Parameter Grid*

### S3.1 Introduction

The main text reports PWA-level matching using two summaries at each PWA's best-fit (l, p, σ) configuration (Section 2.5): the per-PWA MAX is the maximum matched-category count across the 50 seeds at that configuration, and the median-based estimate is the median across the same 50 seeds, the overall effect of that configuration collapsed across random seeds. Because the median is taken across seeds at a fixed configuration, it carries no per-seed re-optimization and is bounded above by the MAX. This supplement presents the detailed statistics underlying the per-PWA MAX result together with its relationship to the median-based estimate. We report this analysis to (i) document the full per-PWA MAX statistics, (ii) quantify how much the median-based estimate differs from the per-PWA MAX result, and (iii) confirm that the principal conclusions hold under both summaries.

### S3.2 Methods

#### S3.2.1 Per-PWA MAX Selection

The per-PWA MAX summary commits each PWA to a single best-fit (l, p, σ) configuration: for each PWA, the configuration whose maximum matched-category count across the 50 seeds is highest is selected by searching the full 4,000-point parameter grid (10 noise levels × 10 modification proportions × 40 layers); the per-PWA MAX is that maximum matched-category count. The median-based estimate, by contrast, is the median of the matched-category count across the same 50 seeds at that committed configuration, summarizing the configuration's typical rather than peak performance across random seeds.

#### S3.2.2 Relationship to the Median-Based Estimate

For the median-based estimate, each PWA's best-fit (l, p, σ) configuration is first fixed (the configuration committed by the per-PWA MAX rule above), and the median of the matched-category count across all 50 seeds at that configuration is taken. This summarizes the typical performance of the committed configuration across random seeds rather than its single best seed, and is bounded above by the per-PWA MAX by construction. A separate selection, the configuration that maximizes the cross-seed median (termed the median-optimal configuration), is examined in Section S3.7 as a probe of parameter-space degeneracy.

#### S3.2.3 Shared Statistical Framework

Both methods are evaluated against the same Monte Carlo baselines established through 50,000 simulations (Supplementary Material S2). Statistical tests include: Shapiro-Wilk normality assessment, chi-square goodness-of-fit, one-sample t-test against baseline, cross-seed one-way ANOVA, Bonferroni correction ($\alpha = 0.05/3 \approx 0.0167$), Cohen’s d and h effect sizes, Wilson confidence intervals, and bootstrap confidence intervals (1,000 iterations).

### S3.3 Comprehensive Statistical Comparison

| Metric | median-based (best-fit (l, p, σ), 2 SD tolerance) | per-PWA MAX (best-fit (l, p, σ), 2 SD tolerance) |
|---|---|---|
| PWAs with ≥6/7 | 155 (55.8%) | 272 (97.8%) ± 7.1 |
| PWAs with =7/7 | 104 (37.4%) | 221 (79.5%) ± 8.1 |
| Mean of best matched-cat count | 5.70 ± 1.25 | 6.77 |
| MC baseline (Real Selection, ≥6/7, 2 SD tolerance) | 34.86% | 34.86% |
| MC baseline (Generated, ≥6/7, 2 SD tolerance) | 25.82% | 25.82% |
| Cross-seed CV | n/a (deterministic at fixed (l, p, σ)) | jackknife SE ±7.1 (≥6/7), ±8.1 (7/7) |
| Algorithmic property | Median across seeds at fixed (l, p, σ) | Per-PWA maximum across seeds at fixed (l, p, σ) |

*Table S12. Comprehensive statistical comparison of the per-PWA MAX result and the median-based estimate on the analysis (N = 278 PWAs, 50 seeds). MC Real Selection baseline: 34.86% (2 SD tolerance, ≥6/7), 62.65% (2 SD tolerance, ≥5/7). MC Generated baseline: 25.82% (2 SD tolerance, ≥6/7), 63.38% (2 SD tolerance, ≥5/7). Both summaries achieve highly significant improvements over baseline; the median-based estimate is the median across seeds at each PWA's best-fit (l, p, σ).*

### S3.4 Paired Per-PWA Comparison

Since both summaries evaluate the same 278 PWAs at each PWA's best-fit (l, p, σ) configuration, a paired comparison quantifies the per-PWA score difference. For each PWA, the per-PWA MAX score is the maximum matched-category count across the 50 seeds at that configuration, while the median-based score is the median of the matched-category count across the same 50 seeds at the same configuration.

| Metric | Value (2 SD tolerance) | Note |
|---|---|---|
| Mean per-PWA MAX score | 6.77 | Per-PWA maximum at best-fit (l, p, σ) |
| Mean median-based score | 5.70 | Median across seeds at best-fit (l, p, σ) |
| Mean difference (median-based − Max-based) | −1.08 | Systematic reduction by construction |
| PWAs: median-based > Max-based | 0 | Never exceeds (median ≤ max) |
| Max HQ rate (2 SD tolerance) | 97.8% | Per-PWA maximum at best-fit (l, p, σ) |
| median-based HQ rate (2 SD tolerance) | 55.8% | Median across seeds at best-fit (l, p, σ) |
| HQ rate difference | 42.1 pp | Cost of robustness |
| Wilcoxon signed-rank test | All non-zero diffs same sign | Significant paired difference (p-value < $10^{-10}$) |

*Table S13. Per-PWA paired comparison between the per-PWA MAX result and the median-based estimate at 2 SD tolerance (N = 278 PWAs). Scores represent the number of matched error categories (range 0–7). The systematic median-based ≤ per-PWA MAX relationship is expected (the median across seeds is bounded above by the maximum) and the Wilcoxon signed-rank test confirms the difference is statistically significant.*

## S3.5 Median-Based Category Distribution

| Categories Matched (Median) | PWAs | Percentage | Cumulative % |
|---|---|---|---|
| 7 (exact) | 104 | 37.4% | 37.4% |
| 6.5 | 4 | 1.4% | 38.8% |
| 6 | 47 | 16.9% | 55.8% |
| 5.5 | 5 | 1.8% | 57.6% |
| 5 | 61 | 21.9% | 79.5% |
| 4.5 | 1 | 0.4% | 79.9% |
| 4 | 44 | 15.8% | 95.7% |
| 3 | 11 | 4.0% | 99.6% |
| 2 | 1 | 0.4% | 100.0% |

*Table S14. Median-based matched-category distribution at 2 SD tolerance (N = 278 PWAs, 50 seeds). The median-based estimate is the median across the 50 seeds at each PWA's best-fit (l, p, σ); because the median across the seeds can fall between integers, half-integer levels appear. High-quality matches (≥ 6 categories): 155 PWAs (55.8%). The mode at 7 categories (37.4%) indicates that, even when each configuration is summarized by its median across seeds rather than its peak, a substantial fraction of PWAs reach perfect matching.*

## S3.6 Discussion

### S3.6.1 Cross-Seed Stability Advantage

### S3.6.2 Relationship between the Per-PWA MAX Result and the Median-Based Estimate

The median-based ≤ per-PWA MAX relationship is expected and follows from the construction of the median: for any sample, the median is bounded above by the maximum. The difference between the per-PWA MAX high-quality rate (97.8%) and the median-based rate (55.8%) reflects the gap between a configuration's peak performance and its median performance across random seeds. The median-based estimate is therefore reported as a companion to the per-PWA MAX result, not as a competing estimate.

### S3.6.3 Per-PWA MAX Result and Median-Based Estimate

Reporting both summaries gives a fuller picture than either alone. The per-PWA MAX result characterizes the matching capacity of the perturbation framework (97.8% high-quality matches at 2 SD tolerance), while the median-based estimate (55.8% high-quality matches, the median across seeds at each PWA's best-fit (l, p, σ)) shows the typical rather than peak performance of that configuration. Together they bracket the per-PWA high-quality matching rate between the median-based 55.8% and the per-PWA MAX 97.8%.

## S3.7 Per-PWA Comparison: Per-PWA MAX vs Median-Optimal Configurations

Figure S1 provides the median-optimal counterpart to main-text Figure 7B. The same six PWAs are shown using the same sampling and the same stacked-bar layout (left bar = PWA observed, right bar = Model output, with the seven response categories stacked in fixed order and consistent color encoding). The difference is the configuration selection algorithm: main-text Figure 7B reports each PWA's per-PWA maximum-matching (l, p, σ) configuration, whereas here the model bar uses each PWA's median-optimal (l, p, σ) configuration (the configuration that maximizes the cross-seed median matched count) evaluated at the single seed that minimizes L2 distance to the observed distribution.

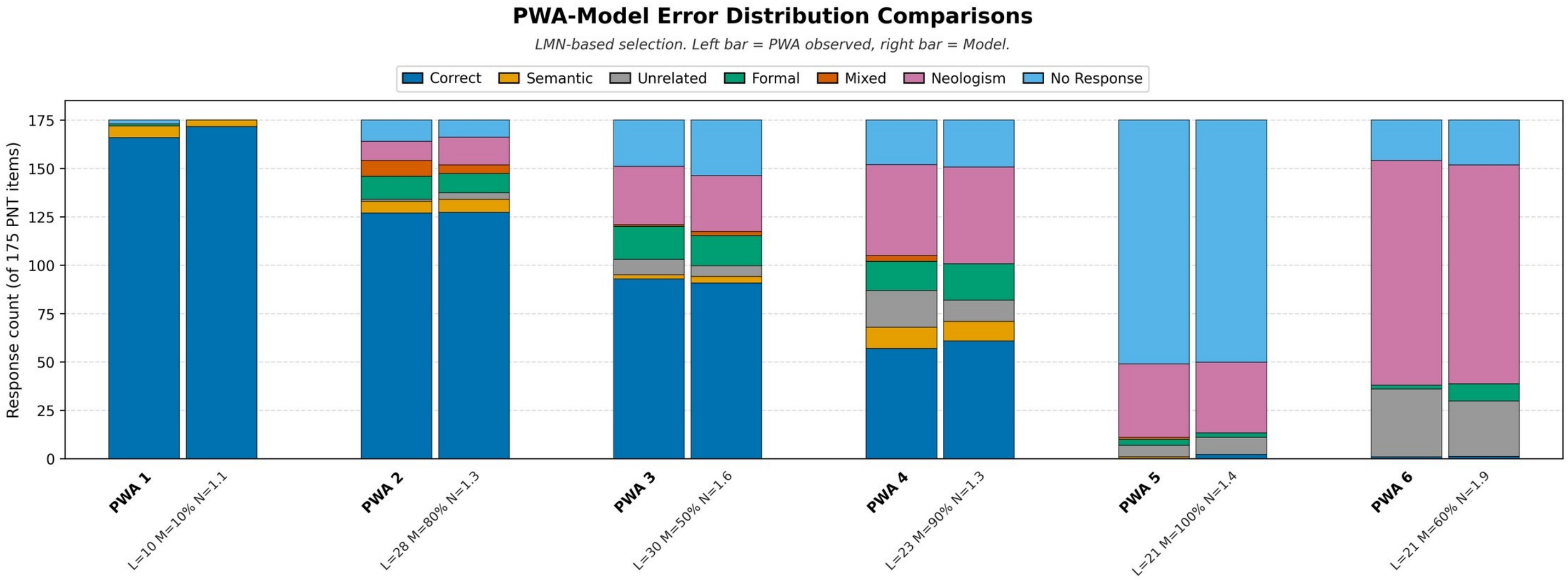

*Figure S1. PWA-Model Error Distribution Comparisons under the median-optimal configuration. For each of the six PWAs (the same selection as main-text Figure 7B, which instead uses the per-PWA maximum-matching configuration), the model bar shows the PWA's median-optimal (l, p, σ) configuration (the configuration that maximizes the cross-seed median matched count) evaluated at the single representative seed that minimizes L2 distance to the observed distribution. Layer (l), proportion (p), and noise (σ) are annotated beneath each pair; the corresponding seed is reported in the manifest accompanying these supplementary materials. The median-optimal and per-PWA MAX configurations generally occupy different regions of the (l, p, σ) space yet reproduce the same observed profiles to within the 2 SD retest-derived tolerance (Table S15), illustrating the degenerate solution structure characterized in Section 3.4 of the main text.*

### S3.7.1 Per-PWA MAX vs Median-Optimal Configuration Divergence

For the six representative PWAs, comparing each PWA's median-optimal configuration (the configuration maximizing the cross-seed median) against its per-PWA MAX best-fit configuration reveals substantial divergence:

| PWA | median-optimal (l, p, σ) | per-PWA MAX (l, p, σ) | Differing parameters |
|---|---|---|---|
| PWA 1 | 10, 10%, 1.1 | 34, 30%, 1.6 | fully different |
| PWA 2 | 28, 80%, 1.3 | 27, 60%, 1.4 | fully different |
| PWA 3 | 30, 50%, 1.6 | 27, 70%, 1.4 | fully different |
| PWA 4 | 23, 90%, 1.3 | 22, 60%, 1.4 | fully different |
| PWA 5 | 21, 100%, 1.4 | 16, 70%, 1.4 | l, p differ |
| PWA 6 | 21, 60%, 1.9 | 13, 90%, 1.8 | fully different |

*Table S15. Per-PWA configuration divergence between the median-optimal and the per-PWA MAX selection. For each PWA the median-optimal configuration, the (l, p, σ) configuration that maximizes the cross-seed median matched count, generally differs from the best-fit configuration that maximizes the per-PWA MAX, yet both reproduce the same PWA error profile to within tolerance (the per-PWA MAX configuration reaches 7/7 for all six PWAs; the median-optimal configuration reaches 7/7 for three and 6/7 for three). This is direct evidence of the degenerate solution structure on the parameter manifold characterized in Section 3.4 of the main text: multiple distinct regions of the manifold can reproduce the same PWA error profile.*

### S3.7.2 L2 Distance Comparison

Although the main text reports the per-PWA MAX result as the primary outcome, the median-optimal configuration, selected to maximize the cross-seed median rather than the per-seed peak, tends to lie at larger L2 distances from the observed distribution on individual PWAs:

| PWA | median-optimal best-seed L2 | per-PWA MAX best-seed L2 |
|---|---|---|
| PWA 1 | 6.66 | 0.54 |
| PWA 2 | 6.84 | 3.10 |
| PWA 3 | 6.38 | 4.87 |
| PWA 4 | 10.62 | 5.66 |
| PWA 5 | 4.30 | 1.31 |
| PWA 6 | 9.99 | 1.86 |
| Mean | 7.46 | 2.89 |

*Table S16. Per-PWA L2 distances between model and observed seven-category distributions for the two selection rules. The per-PWA MAX best-seed mean L2 (2.89) is substantially smaller than the median-optimal mean (7.46), as expected: the per-PWA MAX rule optimizes each PWA's per-seed fit, whereas the median-optimal rule trades per-seed fit for cross-seed robustness. The per-PWA MAX configuration reaches a 7/7 match for all six PWAs and the median-optimal configuration reaches at least 6/7, indicating that both configurations fall within the retest-derived tolerance even though they occupy distinct regions of the parameter manifold.*

### S3.7.3 Median-Optimal Configurations in Parameter Space

Figure S2 is the median-optimal counterpart to main-text Figure 7A. Whereas Figure 7A shows each PWA's best-fit (per-PWA MAX) configuration, Figure S2 plots the median-optimal configuration (the configuration with the highest cross-seed median matched count) for each PWA reproduced in at least six of seven categories under that rule. The two figures share axes and scale, so the contrast is direct: for most PWAs the median-optimal configuration occupies a different region of the parameter space than the best-fit configuration, while still reproducing the error profile, illustrating the manifold degeneracy of Section 3.4.

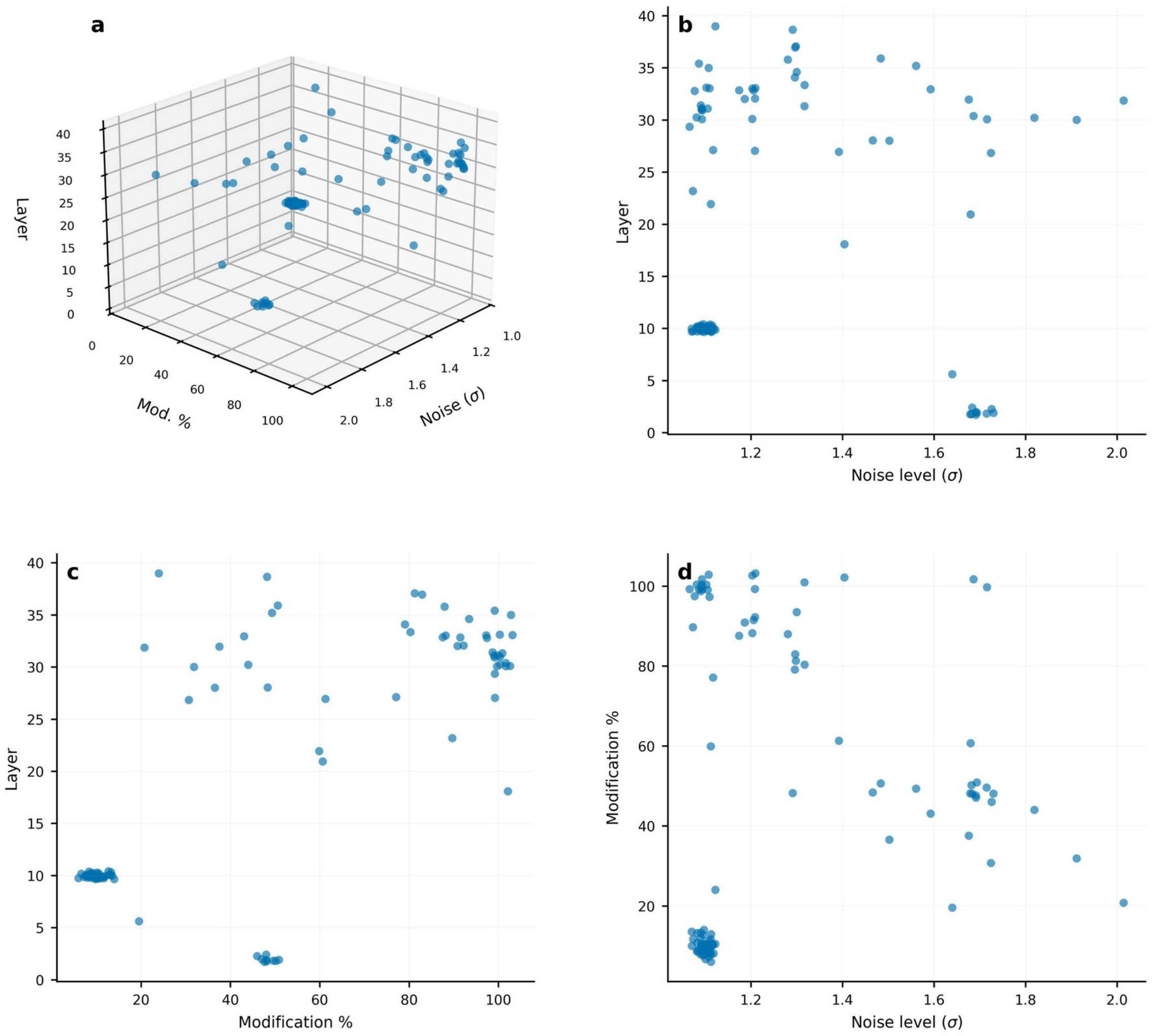

*Figure S2. Median-optimal perturbation configurations reproducing individual PWA error profiles, the counterpart to main-text Figure 7A under the median-optimal selection rule. For each PWA whose median-optimal configuration (the configuration with the highest cross-seed median matched count, ties broken by the cross-seed mean) reproduces the profile in at least six of seven categories (165 PWAs, 59.4%), the marker shows that configuration in the (l, p, σ) parameter space. (a) the joint three-dimensional distribution; (b–d) its two-dimensional projections: (b) layer versus noise, (c) layer versus modification, and (d) modification versus noise. Because each PWA is represented by one selected configuration, the markers fall on discrete grid nodes.*

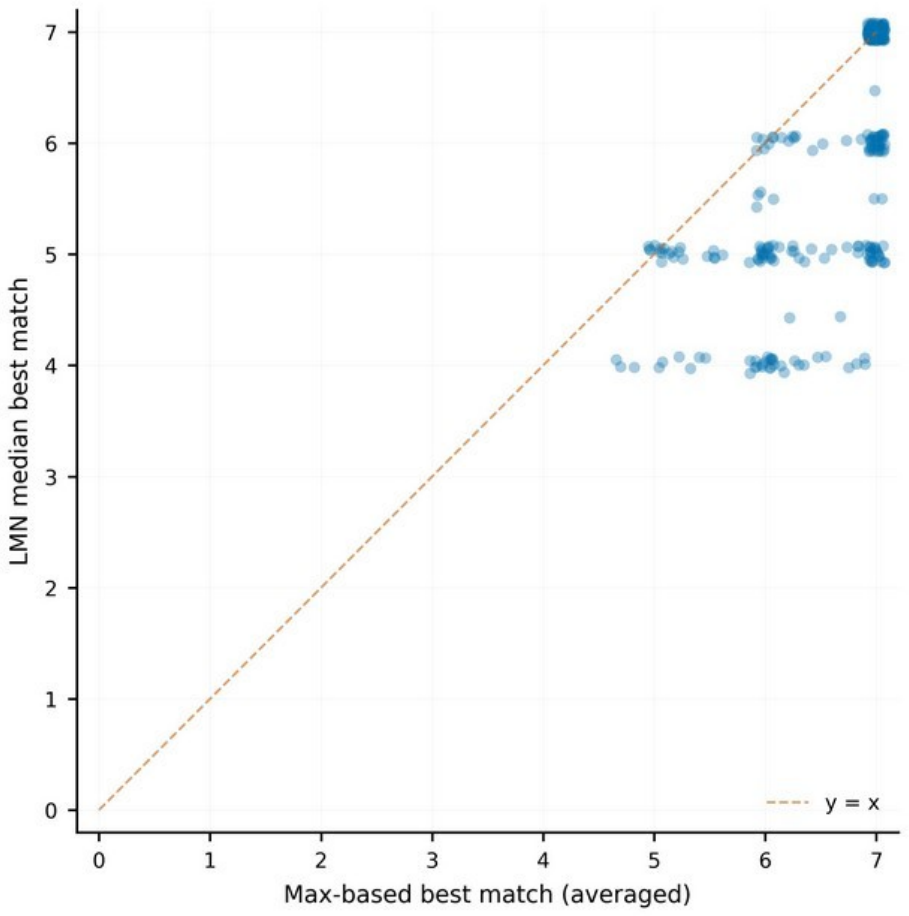

*Figure S3: Per-PWA comparison of the per-PWA MAX and median-based matching scores (N = 278 PWAs, 50 seeds). Each marker is one PWA, plotted by its per-PWA MAX best match (x-axis) against its median-based score (y-axis); the dashed line is y = x. All markers lie on or below the diagonal, showing that the median-based estimate is uniformly less than or equal to the per-PWA MAX: the median across seeds is bounded above by the maximum.*

## S4 — Classification Methodology Validation

*DeBERTa Classifier Architecture, Expert Validation, and Distribution Alignment Analysis*

### S4.1 Background and Rationale

Valid cross-population comparison between large language model (LLM) perturbation errors and PWA naming errors requires methodological consistency in classification. If different classification methods are applied to LLM data versus PWA data, observed similarities or differences may reflect classification artifacts rather than genuine computational parallels. This supplement documents comprehensive validation of three classification approaches to establish the optimal methodology for the main analysis.

Three classifiers were evaluated: (1) Legacy rule-based classifier using Double Metaphone phonetic encoding with orthographic criteria, (2) Sophie rule-based classifier using CMU dictionary phoneme analysis with stricter thresholds, and (3) feature-augmented neural classifier built on the DeBERTa-v3-small backbone (He et al., 2021). Each was evaluated against expert speech-language pathology (SLP) annotations on 3,160 LLM inference records and compared to PWA data distributions from 156 PWAs. The classifier validation in this supplement is independent of the perturbation seed pool used in the main analysis, since classifier training and evaluation use a fixed set of expert-annotated naming responses.

### S4.2 Evaluation Framework

Three complementary metric types were employed. **Distribution-level metrics** measured alignment between classifier output distributions and PWA data distributions (Pearson correlation, Spearman correlation, cosine similarity, mean absolute error (MAE), root-mean-square error (RMSE), Kullback–Leibler (KL) divergence, Jensen-Shannon (JS) divergence, total variation distance). **Item-level metrics** measured agreement between classifier predictions and expert annotations on individual records (accuracy, Cohen's Kappa). Kappa interpretation follows Landis and Koch (1977): 0.81–1.00 Almost Perfect, 0.61–0.80 Substantial, 0.41–0.60 Moderate. **Per-category metrics** identified category-specific strengths and weaknesses (precision, recall, F1-score per category, macro-averaged and support-weighted F1).

### S4.3 Results

#### S4.3.1 Distribution-Level Alignment

| Metric | Legacy | Sophie | DeBERTa | Expert |
|---|---|---|---|---|
| Pearson r | 0.6823 | 0.7712 | 0.9757 | 0.9779 |
| Spearman r | 0.1905 | 0.2857 | 0.9048 | 0.9524 |
| Cosine similarity | 0.8262 | 0.8705 | 0.9626 | 0.9658 |
| MAE (%) | 8.62 | 7.72 | 4.53 | 4.49 |
| RMSE (%) | 11.04 | 9.66 | 6.16 | 5.94 |
| JS divergence | 0.0454 | 0.0454 | 0.0000 | 0.0000 |
| Total variation | 0.3450 | 0.3089 | 0.1810 | 0.1795 |

*Table S17. Distribution-level metrics comparing classifier outputs against PWA data. DeBERTa achieves near-identical alignment to expert annotations across all metrics. Expert annotations shown as reference ceiling.*

#### S4.3.2 Item-Level Agreement

| Classifier | Accuracy | Cohen's Kappa | Interpretation |
|---|---|---|---|
| Legacy rule-based | 66.30% | 0.5831 | Moderate |
| Sophie rule-based | 73.48% | 0.6712 | Substantial |
| DeBERTa neural | 89.21% | 0.8655 | Almost Perfect |

*Table S18. Item-level agreement with expert SLP annotations on 3,160 records. Cohen's Kappa interpretation follows Landis and Koch (1977). DeBERTa achieves Almost Perfect agreement, substantially exceeding both rule-based alternatives.*

#### S4.3.3 Per-Category F1 Scores

| Category | Legacy F1 | Sophie F1 | DeBERTa F1 | Support | DeBERTa Gain |
|---|---|---|---|---|---|
| Correct | 0.983 | 0.988 | 0.993 | 1,154 | +0.005 |
| Semantic | 0.612 | 0.717 | 0.888 | 157 | +0.171 |
| Unrelated | 0.460 | 0.557 | 0.842 | 311 | +0.285 |
| Formal | 0.464 | 0.663 | 0.829 | 304 | +0.166 |
| Nonword | 0.501 | 0.631 | 0.783 | 440 | +0.152 |
| Mixed | 0.471 | 0.574 | 0.821 | 88 | +0.247 |
| Neologism | 0.656 | 0.756 | 0.826 | 341 | +0.070 |
| No Response | 0.000 | 0.000 | 0.875 | 365 | +0.875 |
| Macro-F1 | 0.518 | 0.611 | 0.857 | — | +0.246 |
| Weighted-F1 | 0.633 | 0.700 | 0.892 | — | +0.192 |

*Table S19. Per-category F1-scores against expert annotations. Support indicates number of instances per category. DeBERTa Gain shows improvement over Sophie. The complete failure of both rule-based classifiers on No Response detection (F1 = 0.000) is the most consequential finding.*

### S4.4 The No Response Detection Gap

The most significant finding is the complete failure of both rule-based classifiers on No Response detection (F1 = 0.000). Confusion matrix analysis reveals systematic misclassification: Legacy classifies 66% of No Response instances as Unrelated and 21% as Semantic; Sophie similarly redistributes No Response primarily to Unrelated (60%) and Semantic (14%). This failure reflects fundamental limitations of rule-based approaches: expert annotators recognize diverse No Response manifestations including circumlocutions, descriptive attempts, meta-linguistic comments, and explicit non-responses, patterns requiring pragmatic and discourse-level analysis that lexical matching rules cannot capture.

The No Response category constitutes 11.6% of expert-annotated LLM responses and 15.3% of PWA clinical data. Rule-based classifiers systematically redistribute this substantial category, inflating Unrelated and Semantic error counts while eliminating No Response detection entirely. This produces a 15.25 percentage point discrepancy between classifier outputs (0%) and PWA data (15.3%) for the No Response category alone.

### S4.5 Category-Wise Distribution Alignment

| Category | Legacy | Sophie | DeBERTa | Expert |
|---|---|---|---|---|
| Correct | 15.48 | 15.10 | 14.65 | 14.24 |
| Semantic | 6.18 | 3.65 | 2.70 | 1.94 |
| Unrelated | 19.83 | 14.26 | 5.46 | 5.52 |
| Formal | 1.65 | 5.47 | 3.10 | 3.42 |
| Nonword | 2.12 | 0.54 | 0.51 | 2.72 |
| Mixed | 0.46 | 1.44 | 0.96 | 1.06 |
| Neologism | 8.03 | 6.07 | 5.37 | 3.29 |
| No Response | 15.25 | 15.25 | 3.45 | 3.70 |
| Sum | 68.98 | 61.78 | 36.20 | 35.89 |

*Table S20. Absolute percentage difference from PWA data distribution per category. Lower values indicate closer alignment. DeBERTa's total discrepancy (36.20) approaches the expert ceiling (35.87) and is roughly half the Legacy (68.98) and Sophie (61.78) discrepancies.*

### S4.6 Impact of Classifier Adoption

#### S4.6.1 Distribution Changes

| Category | Legacy % | DeBERTa % | Change | PWA % |
|---|---|---|---|---|
| Correct | 35.3 | 36.1 | +0.8 | 50.8 |
| Semantic | 9.2 | 5.7 | −3.5 | 3.0 |
| Unrelated | 24.1 | 9.8 | −14.3 | 4.3 |
| Formal | 4.6 | 9.3 | +4.7 | 6.2 |
| Nonword | 9.1 | 11.7 | +2.6 | 11.2 |
| Mixed | 2.2 | 2.7 | +0.5 | 1.7 |
| Neologism | 15.5 | 12.9 | −2.6 | 7.5 |
| No Response | 0.0 | 11.8 | +11.8 | 15.3 |

*Table S21. Expected distribution shift from Legacy to DeBERTa classification. Change column shows percentage point difference. PWA column shows target distribution for reference. The emergence of No Response detection (+11.8 pp) is the most consequential change.*

### S4.6.2 Key Impact Areas

**No Response detection:** The most significant change is the emergence of the No Response category (0% to 11.8%), enabling comparison of complete naming-failure patterns between LLM perturbation conditions and PWA profiles. This is essential for matching PWAs with substantial naming-failure rates (mean No Response counts as high as 43.8 out of 175 in the highest-severity band of the cohort), which are characterized by high rates of naming failure on the PNT.

**Unrelated error reduction:** Unrelated errors decrease substantially (24.1% to 9.8%), moving closer to PWA levels (4.3%). This reflects correct reclassification of responses previously misidentified as Unrelated when they were actually No Response or other error types.

**Formal error increase:** Formal errors increase (4.6% to 9.3%), now closer to PWA levels (6.2%), as DeBERTa's phoneme-based detection identifies formal relationships that Double Metaphone encoding missed.

### S4.7 DeBERTa Classifier Architecture

The feature-augmented neural classifier was built on the DeBERTa-v3-small backbone (He et al., 2021) and fine-tuned on 3,160 expert-annotated naming responses. The classifier receives feature-augmented input combining target-response word pairs with six computed binary features: exact match (including singular/plural variants), phonological relatedness, onset phoneme match, word-final phoneme match, lexicality (presence in CMU pronunciation dictionary), and response length flag. This feature augmentation provides explicit phonological and lexical signals that guide classification decisions, supplementing the contextual representations learned by the transformer encoder.

The classifier achieves F1 = 0.875 for No Response detection, a category where rule-based methods completely fail (F1 = 0.000). This capability is essential because No Response errors constitute 11.6% of expert-annotated data and 15.3% of PWA clinical data. Distribution-level validation confirmed that classifier outputs closely match PWA data patterns (Pearson r = 0.9757, Jensen-Shannon divergence = 0.0000), approaching the alignment observed between expert annotations and PWA data (r = 0.9779).

### S4.8 Conclusion

Comprehensive evaluation establishes the feature-augmented neural classifier as the optimal approach for this study, achieving Almost Perfect agreement with expert annotations (Kappa = 0.8655), near-identical distribution alignment with PWA data (r = 0.9757, JS = 0.0000), and successful No Response detection (F1 = 0.875). Rule-based alternatives show only Moderate to Substantial agreement (Kappa = 0.58–0.67) and completely fail on No Response detection, introducing systematic biases that would compromise cross-population comparisons. The adoption of this classifier ensures that observed similarities between LLM perturbation error profiles and PWA naming error profiles reflect genuine computational parallels rather than classification artifacts.

## S5 — Perturbation Locus Sensitivity (LU Control)

### S5.1 Introduction

The main text reports results from the complete-unit (CU) perturbation configuration, in which isotropic Gaussian weight noise is applied uniformly across all 5,120 language-modeling-space units at the target transformer layer. This supplement reports a complementary control analysis in which perturbation is restricted to a Localizer-defined subset of equally-sized language units (language unit, LU, configuration), with the subset computed by identifying the units whose activations were most differentially engaged during the PNT naming task. The LU control is a test of whether the PWA-aggregate matching reported in the main text is specifically attributable to perturbation of localizer-identified units, or whether perturbation of any equally-sized random subset of the 5,120 language units produces equivalent results.

### S5.2 Methods

We constructed the LU configuration by running each of the 175 PNT images through the unperturbed model and recording activations at each transformer layer's LM space; we then computed the per-unit differential engagement across images using a one-vs-rest contrast and ranked units by engagement magnitude. The LU configuration applies perturbation only to the top-ranked subset (size matched to the CU control by selecting the same number of units in expectation under the CU's modification proportion parameter at each (l, p, σ) condition). All other aspects of the perturbation protocol (target layer indexing, noise distribution, modification proportion, classification pipeline, and matching algorithm) are identical to the main-text analyses. We ran the LU analysis on a 20-seed shared subsample of the main analysis (seeds 53021–53028, 53031–53034, 53041–53048, 53052, 53053) using the median-based and the 2 SD retest-derived tolerance.

### S5.3 Results

On the shared 20-seed subsample at 2 SD tolerance, the LU configuration yielded PWA-aggregate high-quality matching of 58.3% (162 of 278 PWAs at ≥6/7 matched categories), compared to 59.4% for the CU configuration on the same 20-seed subsample (Δ = −1.08 percentage points). Perfect 7-category matching was 36.7% for LU versus 37.4% for CU (Δ = −0.72 percentage points). Both differences fall well within seed-level variability (cross-seed CV ≈ 1.5%) and within the category-level tolerance ranges (15.54 responses for correct, 23.16 for neologism). The median-based parameter selection (l, p, σ) was identical between LU and CU for 264 of 278 PWAs (95.0%), with the remaining 14 PWAs (5.0%) showing minor parameter shifts (Δ layer ≤ 4 or Δ proportion ≤ 10%) that did not alter the matched-category count. These results indicate that the PWA-aggregate matching reported in the main text does not depend on the CU configuration choice; perturbation of either the full LM space or a localizer-defined equally-sized subset produces equivalent matching at the PWA-aggregate level.

### S5.4 Discussion

The LU control rules out the possibility that the main-text matching is specifically attributable to perturbation of localizer-identified units; equivalently, it confirms that perturbation of equally-sized random subsets of the LM space produces equivalent PWA-aggregate matching. This is consistent with the manifold view advanced in Section 3.4: the perturbation manifold is shaped by the three control parameters (l, p, σ) rather than by the identity of the specific units perturbed. We emphasize that this control is at the PWA-aggregate level; finer-grained analyses (per-PWA, per-condition, layer-wise dissociation between LU and CU) are beyond the scope of the present report and are deferred to future work that addresses unit-level structure directly. The LU control is methodological: it certifies that the CU choice in the main text does not artificially inflate matching rates, not that the LU and CU configurations produce identical output at the per-PWA or per-condition level.

## S6 — Phonological Re-scoring of the Error Taxonomy

The classifier’s phonological-relatedness feature is a heuristic that, for strings absent from its pronunciation dictionary, can reduce to an orthographic comparison, and its lexicality test does not exclude proper nouns. To establish that the error taxonomy is robust to this operational choice, we re-score the picture-naming responses using an explicit implementation of the standard Philadelphia Naming Test phonological-similarity criterion (Roach et al., 1996), a proper-noun-excluding lexicality test (WordNet), and grapheme-to-phoneme transcription for out-of-vocabulary strings. The re-scoring operates on the existing per-response classifier outputs of the fifty analysis seeds (36,991,481 responses, comprising all 185 PNT images per condition rather than the 158 baseline-correct subset used for patient matching) and adds no new model inference; it re-examines only the formal, mixed, nonword, and neologism categories, on two axes (whether the response is a real word and whether it is phonologically related to the target) and leaves the correct, semantic, unrelated, and no-response categories unchanged. This analysis refines the description of the model’s error structure and does not affect the main patient-matching result, which uses the merged nonword scheme and is unchanged (S6.3).

### S6.1 Method

Each response is reduced to a (target, model-output) pair, where the output is the first whitespace-delimited token. Words present in the CMU Pronouncing Dictionary are transcribed from it; out-of-vocabulary strings are transcribed by a grapheme-to-phoneme model (g2p_en). Because the Philadelphia Naming Test criterion excludes plural morphemes, a trailing /s/ or /z/ is removed before comparison, but only when the de-pluralized spelling is itself a dictionary word.

A target and response are judged phonologically related if any of the following holds, after plural morphemes are removed: (i) they share the initial phoneme, the final phoneme, or the primary-stressed vowel; (ii) they share two or more phonemes at any position, counting consonants and stressed vowels but excluding unstressed vowels; or (iii) they share one or more phonemes at a corresponding position (Roach et al., 1996). The condition satisfied is recorded; the condition number indicates which rule matched, not the strength of the overlap, which is measured separately by target-phoneme coverage below. Condition (iii) is implemented as a match at the same left-aligned position, an approximation of the guide’s corresponding-position rule, which is properly defined over syllabified transcriptions. The criterion reproduces every judgment in the worked examples of the scoring guide.

Lexicality is decided by WordNet membership, which indexes content words and excludes proper nouns; this corrects the original classifier, under which the proper noun “Hampe” counted as a real word. Given the classifier category, the real-word test, and the phonological-relatedness verdict, each response is mapped to a re-scored label by the fixed rule in Table S22.

*Table S22. Deterministic re-scoring rule.*

| Classifier category | Real word? | Phonologically related? | Re-scored label |
|---|---|---|---|
| nonword / neologism | yes | — | misfiled_real_word |
| nonword / neologism | no | yes | phon_related_nonword |
| nonword / neologism | no | no | abstruse_neologism |
| formal | yes | yes | formal_genuine |
| formal | yes | no | formal_no_phonsim |
| formal | no | yes | formal_to_phon_related_nonwo |
| | | | rd |
| formal | no | no | formal_to_abstruse |
| mixed | yes | yes | mixed_phonsim_holds |
| mixed | yes | no | mixed_to_semantic |
| mixed | no | yes | mixed_to_phon_related_nonwor d |
| mixed | no | no | mixed_to_abstruse |

The depth of phonological relatedness is quantified by target-phoneme coverage: the ordered proportion of the target's phonemes that reappear, in sequence, in the response, a value in [0, 1]. High coverage indicates a genuine phonological rendering of the target; low coverage indicates a truncated or degraded semantically related word overlapping the target only incidentally. The 0.5 and 0.67 cut points below are reporting conventions adopted here to separate partial from substantial phonological overlap; they are not diagnostic criteria. Three orthogonal confound flags (format artifact; function word, a closed-class item WordNet does not index; and target prefix, a response that is a proper initial substring of the target) annotate but do not change the label; a response is clean if it carries none of these.

## S6.2 Results

*Table S23. Response category distribution over all 36,991,481 responses.*

| Response category | Count | % of responses |
|---|---|---|
| Correct | 14,741,086 | 39.85% |
| No response | 10,046,802 | 27.16% |
| Neologism (abstruse-type bin) | 5,877,982 | 15.89% |
| Unrelated | 2,384,522 | 6.45% |
| Nonword (phonologically-related bin) | 1,943,112 | 5.25% |
| Semantic | 864,727 | 2.34% |
| Formal | 780,613 | 2.11% |
| Mixed | 352,637 | 0.95% |
| Total | 36,991,481 | 100% |

*Table S24. Re-scored label counts (formal, mixed, nonword, and neologism re-scored).*

| Re-scored label | Count | % of resp. |
|---|---|---|
| abstruse_neologism | 4,183,685 | 11.31% |
| phon_related_nonword | 2,851,312 | 7.71% |
| misfiled_real_word | 724,852 | 1.96% |
| formal_genuine | 494,479 | 1.34% |
| mixed_phonsim_holds | 243,934 | 0.66% |
| formal_to_phon_related_nonword | 173,733 | 0.47% |
| mixed_to_semantic | 82,164 | 0.22% |
| formal_no_phonsim | 65,615 | 0.18% |
| unphonemizable | 61,247 | 0.17% |
| formal_to_abstruse | 46,785 | 0.13% |
| mixed_to_phon_related_nonword | 23,985 | 0.06% |
| mixed_to_abstruse | 2,553 | 0.01% |

*Table S25. Confound decomposition of the phonologically-related-nonword total.*

| Component | Count | % of resp. |
|---|---|---|
| Format artifact | 10,529 | 0.03% |
| Function word | 22,308 | 0.06% |
| Target-word truncation | 192,107 | 0.52% |
| Clean phonologically-related nonword | 2,824,086 | 7.63% |
| Total (permissive criterion) | 3,049,030 | 8.24% |

*Table S26. Target-phoneme coverage of clean phonologically-related nonwords (weighted median 0.333).*

| Coverage bin | Count | % of clean | Cumulative ≥ |
|---|---|---|---|
| [0.0, 0.1) | 29,403 | 1.0% | 100.0% |
| [0.1, 0.2) | 249,763 | 8.8% | 99.0% |
| [0.2, 0.3) | 724,390 | 25.7% | 90.1% |
| [0.3, 0.4) | 584,189 | 20.7% | 64.5% |
| [0.4, 0.5) | 233,592 | 8.3% | 43.8% |
| [0.5, 0.6) | 418,105 | 14.8% | 35.5% |
| [0.6, 0.7) | 302,966 | 10.7% | 20.7% |
| [0.7, 0.8) | 106,204 | 3.8% | 10.0% |
| [0.8, 0.9) | 37,520 | 1.3% | 6.2% |
| [0.9, 1.0] | 137,954 | 4.9% | 4.9% |

About 35.5% of clean phonologically-related nonwords preserve at least half of the target's phonemes (2.71% of all responses) and 10.0% at least two thirds (0.76%). Genuine target-phonological neologisms (spoon → "spooon") therefore occur at roughly twice the genuine formal rate, but are a minority of what the permissive criterion counts; the low-coverage majority are truncated or degraded semantically related words.

*Table S27. Phonologically-related fraction of non-lexical responses by noise level (abstruse fraction is the complement).*

| Noise level | Phon-related fraction | Abstruse fraction |
|---|---|---|
| 1.1 | 0.522 | 0.478 |
| 1.2 | 0.500 | 0.500 |
| 1.3 | 0.482 | 0.518 |
| 1.4 | 0.475 | 0.525 |
| 1.5 | 0.472 | 0.528 |
| 1.6 | 0.449 | 0.551 |
| 1.7 | 0.424 | 0.576 |
| 1.8 | 0.405 | 0.595 |
| 1.9 | 0.393 | 0.607 |
| 2.0 | 0.385 | 0.615 |

Genuine formal errors are near floor (1.34% of responses, 1.07% after removing target-word truncations); the phonologically-related fraction of non-lexical responses falls monotonically as perturbation increases, so abstruse forms grow faster under heavier noise. Of mixed responses that satisfy the phonological criterion, 99.6% rest on a single shared boundary phoneme, and the remainder re-score as semantic substitutions.

### S6.3 Scope

This re-scoring characterizes the model's error structure at the level of the individual response. It does not enter the per-patient best-fit matching pipeline, whose inputs (the merged seven-category distributions) and outputs are identical to those of the main analysis. The headline matching result is therefore unchanged.